%% file: arxiv.tex
\documentclass{article}[10pt,a4paper,twoside]

\usepackage[english]{babel}

\usepackage[letterpaper,top=2cm,bottom=2cm,left=3cm,right=3cm,marginparwidth=1.75cm]{geometry}
\usepackage{xcolor}
\usepackage{hyperref}
\hypersetup{colorlinks,linkcolor=purple, citecolor=blue,urlcolor=blue}
\usepackage[utf8]{inputenc}
\usepackage{amsmath}
\usepackage{amssymb}
\usepackage{bm}
\usepackage{graphicx}

\newcommand{\rebuttal}[1]{\ifnum\debug=1{\color{blue}#1}\else #1\fi}

\def\debug{0}


\title{\textbf{Analytic theory of dropout regularization} 
}
\author{Francesco Mori$^{1,\mathsection}$, Francesca Mignacco$^{2,3}$\\\\
{\small
  $^{1}$Rudolf Peierls Centre for Theoretical Physics, University of Oxford, Oxford OX1 3PU, United Kingdom } \\
  {\small
  $^2$Graduate Center, City University of New York, New York, NY 10016, USA}\\
{\small  $^3$Joseph Henry Laboratories of Physics, Princeton University, Princeton, NJ 08544, USA}\\
{\small $^\mathsection$\texttt{francesco.mori@physics.ox.ac.uk}
}
}

\date{}

\begin{document}
\maketitle

\begin{abstract}
Dropout is a regularization technique widely used in training artificial neural networks to mitigate overfitting. It consists of dynamically deactivating subsets of the network during training to promote more robust representations. Despite its widespread adoption, dropout rates are often selected heuristically, and theoretical explanations of its success remain sparse. Here, we analytically study dropout in two-layer neural networks trained with online stochastic gradient descent. In the high-dimensional limit, we derive a set of ordinary differential equations that fully characterize the evolution of the network during training and capture the effects of dropout. We obtain a number of exact results describing the generalization error and the optimal dropout probability at short, intermediate, and long training times. Our analysis shows that dropout reduces detrimental correlations between hidden nodes, mitigates the impact of label noise, and that the optimal dropout probability increases with the level of noise in the data. Our results are validated by extensive numerical simulations.
\end{abstract}

\maketitle
\section{\label{sec:intro}Introduction}
Dropout is a regularization technique introduced in Ref.~\cite{hinton2012improving,srivastava2014dropout} to mitigate overfitting in artificial neural networks. The key idea is to randomly deactivate a subset of neurons at each training step, effectively training an ensemble of subnetworks with shared weights. Since each neuron can either be active or inactive, the total number of subnetworks grows exponentially with the number of hidden neurons. At testing time, all neurons are used, with their weights appropriately rescaled, leading to an implicit averaging of these subnetworks. This averaging reduces overfitting, as the different nodes are more independent due to dropout, and improves generalization across tasks. Additionally, because not all nodes are active simultaneously, dropout is believed to reduce detrimental co-adaptations and encourage independent representations \cite{hinton2012improving}.

More than a decade after its introduction, dropout remains a widely used technique in training state-of-the-art models \cite{salehin2023review}, with several generalizations proposed. One such variant is DropConnect \cite{wan2013regularization}, which applies dropout at the level of individual weights rather than entire nodes. While early studies \cite{hinton2012improving,srivastava2014dropout} suggested an optimal activation rate between 0.5 and 0.8, constant over training, more recent work has advocated for adaptive schedules \cite{morerio2017curriculum,liu2023dropout}. 

Understanding the effect of dropout on learning has motivated a series of theoretical results. Several papers focused on the worst-case scenario and derived generalization bounds \cite{pmlr-v80-mou18a,arora2021dropout,pmlr-v28-wan13,zhai2018adaptive}. The early work \cite{NIPS2013_71f6278d} formalized the interpretation of dropout as a stochastic regularizer that approximates model averaging in both linear and nonlinear networks. Implicit regularization induced by training with dropout has been studied in generalized linear models \cite{wager2013dropout}, two-layer linear networks \cite{mianjy2018implicit} and two-layer nonlinear networks \cite{zhang2024implicit}. In random deep neural networks, Ref.~\cite{schoenholz2017deep} has shown that dropout destroys the order-to-chaos critical point, resulting in an upper bound on trainable neural depth.

In contrast to these prior approaches, in this paper we focus on the typical-case scenario and derive an exact analytic characterization of online feature-learning dynamics in the presence of dropout regularization. To this end, we leverage tools from statistical physics to obtain closed-form equations for the dropout dynamics in two-layer networks \cite{biehl1995learning,saad1995PRL,saad1995PRE,goldt2019dynamics} within a prototypical teacher-student framework \cite{gardner1989three,PhysRevA.45.6056,engel2001statistical}. This theoretical framework allows us to address several open questions about the mechanisms driving the performance improvement induced by dropout. In particular, we clarify the benefits of dropout in the presence of dataset noise, identify optimal dropout rates, and analyze its impact on the different phases of training.

The remainder of the paper is organized as follows. In Section~\ref{sec:theory}, we introduce the model and provide a summary of our main results. In Section~\ref{sec:results}, we present a detailed explanation of our derivations and findings. In particular, Section~\ref{sec:numeric} offers a numerical analysis of the effects of dropout on the learning dynamics. In Section~\ref{sec:early_times}, we provide an analytical characterization of the early-time dynamics. Section~\ref{sec:uspecialized} focuses on the intermediate-time regime, where the network nodes approximately learn the same features. In Section~\ref{sec:linear_stability}, we analyze the long-time regime, where the nodes specialize to represent different aspects of the task, and provide an analytical prediction for how dropout affects this specialization transition. Section~\ref{sec:long_time} offers further details on the long-time behavior, based on numerical analysis. Finally, we conclude in Section~\ref{sec:discussion} with a summary and future perspectives. Some technical details are relegated to the appendix and to supplementary \textit{Mathematica} notebooks available at~\cite{dropout_notebooks}.

\section{\label{sec:theory}Model and summary of the main results}

In this paper, we consider supervised learning within the teacher-student framework~\cite{gardner1989three,PhysRevA.45.6056,engel2001statistical} for two-layer neural networks. The teacher-student setup is standard in the statistical physics of neural networks and consists of a \emph{student} network that is trained on data with labels generated by passing random inputs through a fixed \emph{teacher} network. In this section, we first introduce our model and then we provide a summary of our main results.

\subsection{Model definition}

\label{sec:model}

\begin{figure}
    \centering
    \includegraphics[width=0.95\linewidth]{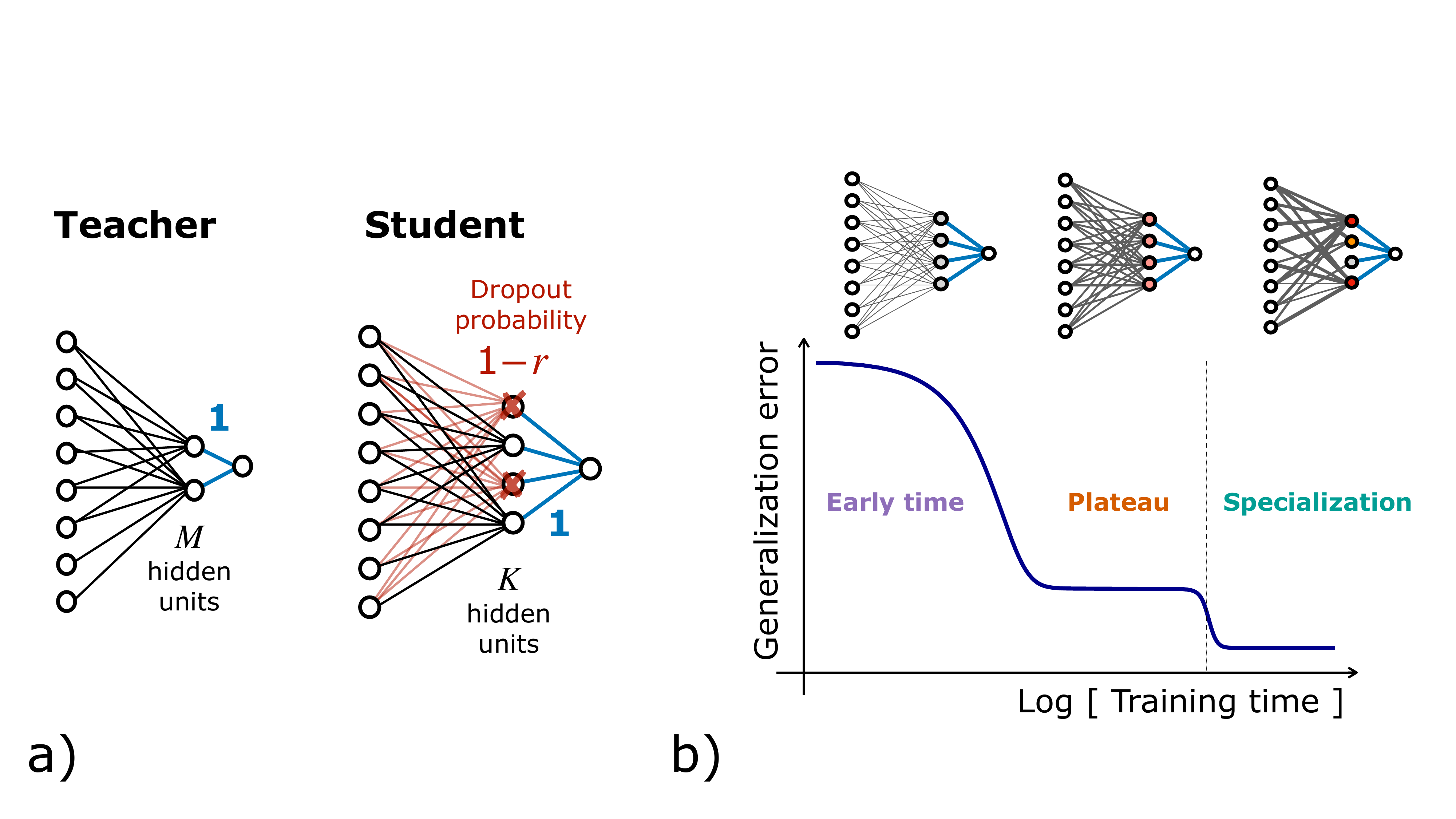}
 \caption{\rebuttal{\textbf{a)} Illustration of the model presented in Section \ref{sec:model}. \textbf{b)} Sketch of the generalization error dynamics highlighting the three phases of learning. 
At early times, the weights remain close to their small-norm initialization. 
During the plateau (unspecialized phase), the hidden units are symmetric, being equally aligned with the teacher units. 
After specialization, this symmetry is broken.}}
    \label{fig:illustration}
\end{figure}

We assume that the input data $x\in \mathbb{R}^N$ are drawn independently as standard Gaussian variables. The corresponding label $y$ is generated via a teacher network $\phi^*(x)$ as
\begin{equation}
    y=\phi^*(x)+\sigma \zeta\,,
\end{equation}
with label noise $\zeta\sim\mathcal{N}(0,1)$ while $\sigma>0$ controls the noise level, and 
\begin{equation}
   \phi^*(x)=\sum_{m=1}^M v^*_m \,g^*\left(\frac{w^*_m  x}{\sqrt{N}}\right)\,,
\end{equation}
where $v^*\in\mathbb{R}^M$, $w^*_m\in\mathbb{R}^{N}$, and $g^*$ denote the fixed teacher weights and activation function.
To learn the dataset, we employ a student network, which after $\mu$ training steps takes the form
\begin{equation}
    \phi^{\mu}(x)=\sum_{i=1}^K \sigma^{\mu}_i v^{\mu}_i g\left(\frac{w^{\mu}_i  x}{\sqrt{N}}\right)\,,
\end{equation}
where $v^{\mu}\in\mathbb{R}^K$ and $w_i^{\mu}\in\mathbb{R}^{N}$ are trainable weights. We have introduced the selection variables 
\begin{equation}
    \sigma_i^{\mu}=\begin{cases}
    1\quad \quad &\text{if node $i$ is active at step $\mu$}\,,\\
    0\quad \quad &\text{otherwise}\,.
    \end{cases}
\end{equation}
We assume that these variables are independently drawn from a Bernoulli distribution with activation probability $0<r\leq 1$, i.e., $\sigma_i^{\mu}\sim \operatorname{Bernoulli}(r)$. These binary variables model the effect of dropout: at each training step $\mu$, each hidden node is active with probability $r$ or dropped out with probability $1 - r$. \rebuttal{A pictorial representation of this model is shown in Figure~\ref{fig:illustration}a.}

We consider training via online stochastic gradient descent (SGD) on the square loss, where each training step involves a single fresh example from the dataset. The update rules are given by
\begin{align}
    w_k^{\mu+1} &= w_k^{\mu} - \frac{\eta}{\sqrt{N}} \sigma_k^{\mu} v_k^{\mu} g'(\lambda_k^{\mu}) \Delta^{\mu} x^{\mu}\,, \\
    v_k^{\mu+1} &= v_k^{\mu} - \frac{\eta_v}{N} \sigma_k^{\mu} g(\lambda_k^{\mu}) \Delta^{\mu}\,,
\end{align}
where $\eta$ and $\eta_v$ are the learning rates for the first and second layers, respectively. Here, the updates are evaluated on an independent data point $\{x^{\mu}, y^{\mu}\}$ freshly drawn at each step. Note that this setup differs from offline (or batch) learning, where training examples are typically reused multiple times. We have defined the error term $\Delta^{\mu}=\hat{y}^{\mu}-y^{\mu}$ and the pre-activations $\lambda_k^{\mu}=w_kx^{\mu}/\sqrt{N}$, for each time step $\mu$. This type of dynamics was previously investigated in Refs.~\cite{saad1995PRL,biehl1995learning,goldt2019dynamics}, where closed equations for low-dimensional order parameters were derived. Here, we would like to extend these results to the case of dropout. 

As customary when using dropout, at testing time the full network $\phi_f(x)$ is used, after a rescaling of the rates by a factor $0 < r_f \leq 1$:
\begin{equation}
    \phi_f(x) = \sum_{i=1}^K r_f\, v^{\mu}_i\, g\left(\frac{w^{\mu}_i x}{\sqrt{N}}\right)\,.
\end{equation}
This rescaling compensates for the reduced activation during training, ensuring consistent output magnitudes between training and testing. For simplicity, in the rest of the paper we consider $r_f = r$. We define the generalization error as
\begin{equation}
    \epsilon_g=\left\langle\frac12 \left[\phi^*(x) - \phi_f(x)\right]^2 \right\rangle\,,
\end{equation}
where the square brackets denote an average over the distribution of the input $x$.

A remarkable feature of this teacher-student model is that it allows for an exact analytical characterization in the thermodynamic (high-dimensional) limit, where the input dimension $N$ and the number of training steps $\mu$ jointly tend to infinity at a fixed ratio $\alpha = \mu / N$. Here, $\alpha$ serves as a continuous version of the training step and is interpreted as the training time. 

Following Ref.~\cite{goldt2019dynamics}, we derive a set of ODEs that describe the training dynamics. To proceed, we introduce a set of low-dimensional quantities, known as \rebuttal{\emph{order parameters}} in statistical physics, that fully characterize the network’s performance in the thermodynamic limit. For our model, the order parameters are
\begin{align}
    Q_{ik}^\mu \equiv \frac{w_i^\mu  \cdot w_k^{\mu}}{N}\,, \quad
    R_{in}^\mu \equiv \frac{w_i^\mu  \cdot w_n^{*}}{N}\,, \quad
    T_{nm} \equiv \frac{w_n^* \cdot  w_m^{*}}{N}\,,
\end{align}
where the symbol $\cdot$ denotes the scalar product and we use the convention that the indices $i, j, k, \ell$ refer to the student network (running from $1$ to $K$), while $m, n$ refer to the teacher (running from $1$ to $M$). Unless stated otherwise, we will assume $T_{nm} = \delta_{nm}$. The order parameters provide physical intuition into the learning dynamics. For instance, $R_{in}$---a.k.a.~the \emph{magnetization}---quantifies the alignment between the $i$-th student node and the $n$-th teacher node. 

In the thermodynamic limit $N,\mu\to \infty$ with $\alpha=\mu/N$ fixed, they satisfy the ODEs
\begin{align}
\label{eq:ODE_dropout_main}
\frac{\mathrm{d} R_{in}}{\mathrm{d} \alpha} = f_{R_{in}}(Q,R,v;\eta,r),  &&
\frac{\mathrm{d} Q_{ik}}{\mathrm{d} \alpha} = f_{Q_{ik}}(Q,R,v;\eta,r),  &&
\frac{\mathrm{d} v_i}{\mathrm{d} \alpha} = f_{v_i}(Q,R,v;\eta,r), 
\end{align}
where the definitions and derivations of $f_{R}$, $f_Q$, and $f_v$ are presented in Appendix~\ref{appendix:derivationODE}.

For simplicity, in what follows we take $g^*(\cdot)=g(\cdot)=\operatorname{erf}\left(\frac{\cdot}{\sqrt{2}}\right)$. For this choice the expressions for the ODEs simplify substantially. In this case, the generalization error can be written as
\begin{align}
\begin{split}
    \epsilon_g &= \frac{r_f^2}{\pi} \sum_{i,k} v_i v_k \arcsin\left(\frac{Q_{ik}}{\sqrt{1 + Q_{ii}} \sqrt{1 + Q_{kk}}}\right)
\\&\quad+ \frac{1}{\pi} \sum_{n,m} v_n^* v_m^* \arcsin\left(\frac{T_{nm}}{\sqrt{1 + T_{nn}} \sqrt{1 + T_{mm}}}\right)\\
&\quad- \frac{2r_f}{\pi} \sum_{i,n} v_i v_n^* \arcsin\left(\frac{R_{in}}{\sqrt{1 + Q_{ii}} \sqrt{1 + T_{nn}}}\right).\label{eq:gen_err}
\end{split}
\end{align}
Note that by setting $r = r_f = 1$, thereby removing dropout, we recover the equations of the teacher-student model studied in Ref.~\cite{goldt2019dynamics}. Our first result is thus the extension of these equations to incorporate the effects of dropout.

Our main focus is on the over-parameterized regime $K > M$, in which the student network has sufficient capacity to reproduce the teacher network exactly and also represent more complex functions. For simplicity, we do not train the second layer, by fixing $v_i=v^*_n=1$ and $\eta_v=0$.  In this setting, the teacher and student networks are known as soft committee machines \cite{saad1995PRE}.

\begin{figure}
    \centering
    \includegraphics[width=0.32\textwidth]{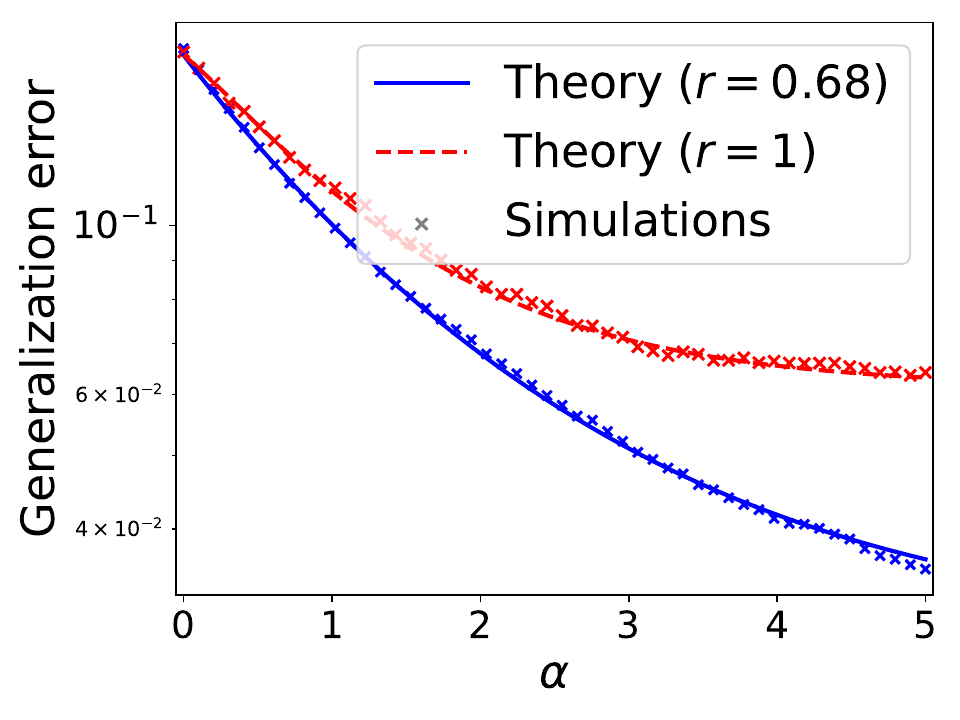} \includegraphics[width=0.32\textwidth]{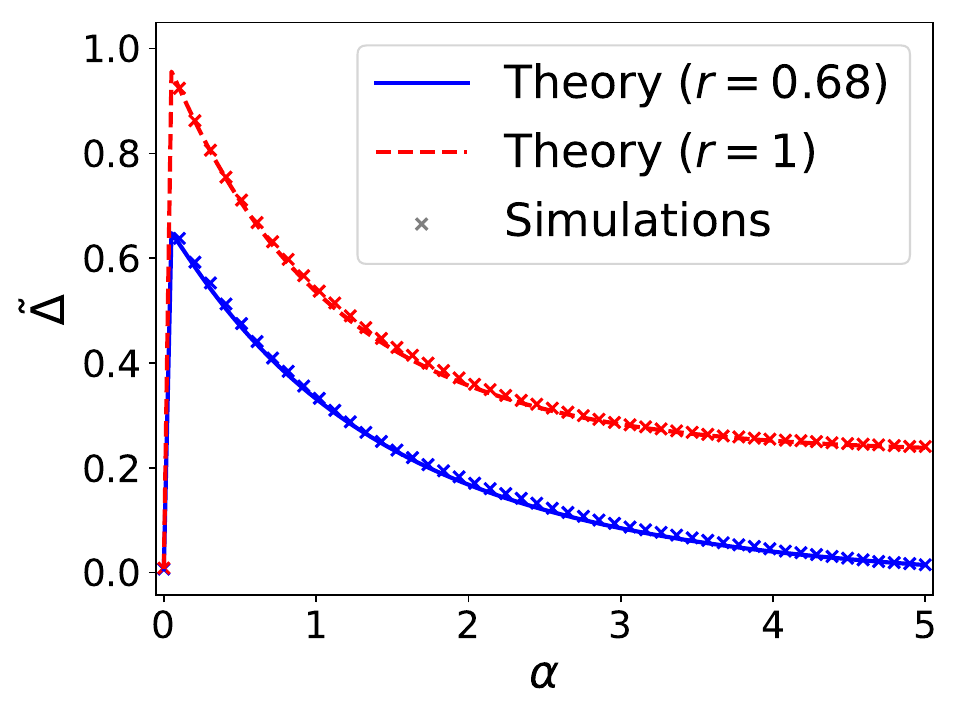}
    \includegraphics[width=0.32\textwidth]{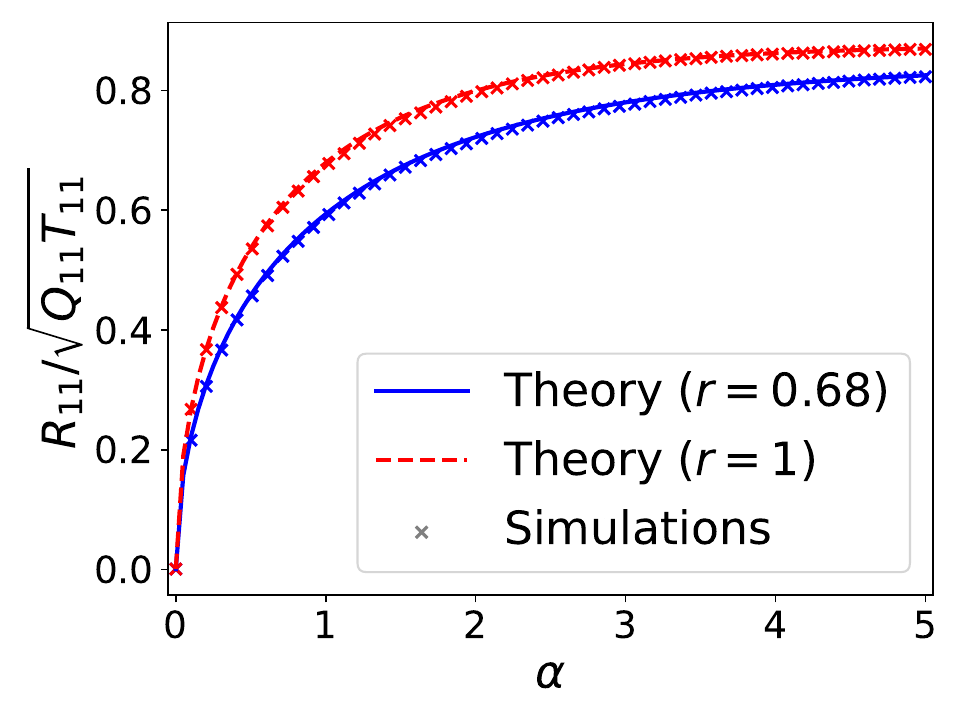}
    \caption{\textbf{Comparison of learning dynamics with and without dropout.} \textbf{Left:} Generalization error vs. training time $\alpha=\mu/N$ with ($r = 0.68$) and without ($r = 1$) dropout. \textbf{Center:} The observable $\tilde{\Delta}=(Q_{12} - R_{11} R_{21}) / \sqrt{Q_{11} Q_{22}}$, which quantifies detrimental correlations between the student’s hidden nodes (see main text for details). \textbf{Right:} Cosine similarity between the student’s first hidden node weight vector $w_1$ and the teacher weight vector $w^*_1$. 
\textbf{Parameters:} $K = 2$, $M = 1$, $\eta = 1$, $\eta_v = 0$, $\alpha = 5$, and $\sigma = 0.3$. The teacher weights $w^*$ are drawn i.i.d.~from $\mathcal{N}(0,1)$; the student weights are initialized i.i.d.~from $\mathcal{N}(0,0.1)$, with $N = 50000$. The second-layer weights are fixed to one. Solid lines show theoretical predictions from the ODEs in Eq.~\eqref{eq:ODE_dropout_main}; crosses mark numerical simulations on a single training trajectory.}%figures generated with dropout_nocontrol_simulations.ipynb
\label{fig:comparison_dropout_nodropout}
\end{figure}

The results of numerically integrating the ODEs~\eqref{eq:ODE_dropout_main} are presented in Fig.~\ref{fig:comparison_dropout_nodropout} for different values of the activation probability $r$. The continuous lines, obtained from the ODEs, show excellent agreement with the numerical simulations (crosses) performed on a single training trajectory with $N = 50000$. This confirms that our theoretical predictions, derived in the thermodynamic limit $N \to \infty$, remain accurate even at finite system size. Interestingly, in the left panel of Fig.~\ref{fig:comparison_dropout_nodropout} we observe that dropout leads to a substantial performance improvement with respect to training without dropout.

\subsection{Summary of the main results}

To aid the reader, we provide a brief summary of the main results. Full derivations are presented in the following sections.
\begin{itemize}
\item \textbf{Analytic theory of dropout:} We develop an analytic framework describing the online SGD learning dynamics with dropout in a prototypical teacher-student model. The dynamics are captured by a set of ODEs, given in Eq.~\eqref{eq:ODE_dropout_main}, which are exact in the high-dimensional limit $N \to \infty$. While these ODEs cannot be solved analytically in general, we first examine their numerical integration and then explore specific limits where analytical progress can be made. \rebuttal{An illustration of the different learning phases under consideration is provided in Figure~\ref{fig:illustration}b.}

\item {\bf Numerical analysis of dropout:} As a first step, we integrate Eq.~\eqref{eq:ODE_dropout_main} numerically to investigate the effects of dropout on the high-dimensional learning dynamics (see Fig.~\ref{fig:comparison_dropout_nodropout}). We observe that, although dropout decreases the cosine similarity between student and teacher nodes, it leads to more independent student representations, resulting in a substantial improvement in performance. In particular, we find that dropout mitigates the negative effects of label noise. Indeed, the optimal activation rate $r^*$ decreases with the variance $\sigma^2$ of the label noise (see Fig.~\ref{fig:optimal_const_dropout}).

\item \textbf{Early-time dynamics:} At the beginning of training, we assume that the student weights are initialized with a small norm and expand the dynamics to leading order in that norm. The evolution equations can be solved exactly for arbitrary $M$, $K$, and $r$. As a result, we compute the optimal activation rate that minimizes the generalization error:
\begin{equation}
\begin{split}
    r^* =\min\left\{1;\frac{12 M}{3M \eta + 9\eta \sigma^2 + \sqrt{3\eta (M + 3\sigma^2)\left[M(32(K - 1) + 3\eta) + 9\eta \sigma^2\right]}}\right\}\,,
    \end{split}
\end{equation}
and matches well with the numerical result shown in Fig.~\ref{fig:optimal_const_dropout}. Interestingly, $r^*$ decreases with increasing noise variance $\sigma^2$, consistent with the intuition that dropout prevents noise fitting. It also decreases with an increasing number of hidden nodes $K$, reflecting the role of dropout in mitigating negative co-adaptations between hidden units. Furthermore, we obtain the phase diagram in Fig.~\ref{fig:phase_digram}, which identifies the regimes in which dropout improves generalization as a function of the problem parameters.

\item \textbf{Unspecialized phase (\rebuttal{analyzed in the} low-$\eta$ limit):} We find that the generalization error quickly reaches a plateau, corresponding to an unspecialized phase in which each student node overlaps equally with all teacher nodes. This unspecialized phase\rebuttal{, observed numerically across different values of the learning rate,} is known to be unstable in the absence of dropout~\cite{saad1995PRE}. The plateau corresponds to a fixed point of the ODEs, which cannot be computed analytically in general. However, in the limit of small learning rate $\eta$, it is possible to derive an analytic expression for this fixed point. In this regime, we obtain the following exact expression for the generalization error (see also Fig.~\ref{fig:low_eta_theory})
\begin{equation}
\begin{split}
    \epsilon_g^0 =\frac{1}{\pi} \left( \frac{M \pi}{6} + 
    K^2 r^2 \arcsin\left( \frac{M R_0^2}{1 + M R_0^2} \right) - 
    2 K M r \arcsin\left( \frac{R_0}{\sqrt{2} \sqrt{1 + M R_0^2}} \right) \right),\label{eq:gen_intro}
    \end{split}
\end{equation}
where 
\begin{equation}
\begin{split}
\frac{1}{R_0
}= 
  \sqrt{ \left( 1 + (K-1)r \right)^2 -M + \sqrt{ M^2 + (K-1)r \left(1 + (K-1)r\right)^2 \left(2 + (K-1)r\right) } } \,.
\end{split}
\end{equation}
The optimal activation rate $r_0^*$ is obtained by minimizing the expression for the generalization error and is shown, together with the corresponding generalization error, in Fig.~\ref{fig:rstar_plateu}. We observe that the activation probability decreases with increasing $K$. Note, however, that in the ODEs~\eqref{eq:ODE_dropout_main}, the dependence on the label noise variance $\sigma^2$ disappears at leading order in $\eta$. This motivates us to compute the first-order correction in $\eta$ to determine how the optimal dropout rate depends on $\sigma^2$.

\item \textbf{Unspecialized phase (\rebuttal{introducing} small-$\eta$ corrections):} To investigate the role of label noise in determining the optimal activation rate, we derive perturbative corrections to the fixed point to leading order in $\eta$. In particular, we show analytically that the optimal activation probability takes the form
\begin{equation}
    r^* \approx r_0^* + \eta \left(a(K, M) + b(K, M) \sigma^2\right)\,,
\end{equation}
where $a(K, M)$ and $b(K, M)$ are coefficients we compute explicitly (see Fig.~\ref{fig:coeff}). Notably, we find that $b(K, M) < 0$ for all values of $K$ and $M$ considered, confirming that the optimal activation probability decreases with increasing label noise.

\item \textbf{Specialization transition:} As anticipated, in the absence of dropout ($r = 1$), the unspecialized phase is known to be linearly unstable~\cite{saad1995PRE}, leading to \rebuttal{specialization} at long times—i.e., each student node aligns preferentially with one of the teacher nodes. Here, we investigate how dropout modifies this specialization transition. By performing a linear stability analysis around the plateau fixed point, we derive the phase diagram shown in Fig.~\ref{fig:phase_digram_special}, valid for $K = M$ and small $\eta$. Interestingly, we find that specialization occurs only for sufficiently large values of $r$ (i.e., low dropout). Below a critical threshold in $r$, the plateau phase becomes stable and the system does not transition to specialization. \rebuttal{Finally, for $K=M$ and small $\eta$, we find that the optimal activation probability is given by 
\begin{equation}
    r^*\approx\min\left[1; \, 1+\eta(a(M)+\sigma^2 b(M))\right]\,,
\end{equation}
where $a(M)$ and $b(M)$ can be computed explicitly (see Fig.~\ref{fig:coeff_specialized}). We find $b(M)<0$ for all values of $M$ considered, confirming that dropout helps to mitigate label noise even in the specialized phase.
}

\end{itemize}
\section{\label{sec:results}Results}

In this section, we begin by discussing general insights obtained from the numerical integration of the ODEs~\eqref{eq:ODE_dropout_main}. We then proceed with an analytical study of these equations, deriving explicit results for the optimal activation rate and the corresponding generalization performance across the different phases of the learning dynamics.

\subsection{Numerical analysis of the dropout dynamics}\label{sec:numeric}

To understand the performance improvement introduced by dropout, it is helpful to examine the evolution of the order parameters over training time (see Fig.~\ref{fig:comparison_dropout_nodropout}). For simplicity, we analyze the case $K=2$ and $M=1$, though the following reasoning extends to scenarios where $K > M$.

During training, the weight vectors $w_1$ and $w_2$ associated with the two student hidden nodes progressively align with the teacher weight vector $w_1^*$. However, due to noise (arising from the randomness in the input $x$, label noise, and random initial conditions), $w_1$ and $w_2$ also acquire an additional random component, perpendicular to $w^*_1$. At any point during training, the weights can be expressed as:  
\begin{equation}
    w_i = R_{i,1} w_1^* + \tilde{w}_i,\label{eq:decomposition}
\end{equation}
where $\tilde{w}_i$ represents the noisy component, with $\tilde{w}_i \cdot w^*_1=0$. Ideally, we seek a large $R_{i,1}$ relative to the norm of $\tilde{w}_i$ and prefer $\tilde{w}_1$ and $\tilde{w}_2$ to be uncorrelated. Indeed, when these noisy components are uncorrelated, their adverse effects on network performance tend to average out. Conversely, if they are correlated, their detrimental effect is compounded. These detrimental correlations can be expressed in terms of the order parameters as
\begin{equation}
   \tilde{\Delta}\equiv \frac{\tilde{w}_1 \cdot  \tilde{w}_2}{\sqrt{w_1 \cdot  w_1 \, w_2 \cdot  w_2}} = \frac{Q_{12} - R_{11} R_{21}}{\sqrt{Q_{11} Q_{22}}}\,.
\end{equation}
This quantity is plotted in the central panel of Fig.~\ref{fig:comparison_dropout_nodropout}. Interestingly, we observe that dropout reduces these correlations, mitigating the interference between student hidden nodes.

We can therefore identify two competing objectives: 1) learning the signal $w_1^*$ effectively (favored by a low dropout rate) and 2) avoiding fitting the noise by promoting independence between $\tilde{w}_1$ and $\tilde{w}_2$ (favored by a higher dropout rate).

The central panel in Figure~\ref{fig:comparison_dropout_nodropout} illustrates that, without dropout, the quantity $\tilde{\Delta}$ is significantly larger compared to the case with dropout. This indicates that the two nodes learn similar noisy components ($\tilde{w}_1 \approx \tilde{w}_2$), as they are simultaneously trained on the same data points, exacerbating the fitting of noise. By contrast, with dropout, the nodes are trained jointly only for a fraction of the time and individually otherwise, promoting more independent representations. The readout layer then averages these representations, partially canceling the noise and leading to improved performance.

Interestingly, dropout reduces the alignment of the weights with the signal $w_1^*$ (right panel in Fig.~\ref{fig:comparison_dropout_nodropout}). This is expected, as dropout effectively reduces the number of training updates of each individual node. Despite the weaker signal representation, the benefit of increased independence between nodes outweighs this drawback, leading to superior generalization.

A key parameter in our setting is the variance $\sigma^2$ of the label noise. When training without dropout, the network is expected to be more prone to fitting this noise, which can degrade performance. In practical applications, dropout is commonly observed to mitigate noise-fitting and promote a more robust representation of the data.

\begin{figure}
    \centering
    \includegraphics[width=0.6\linewidth]{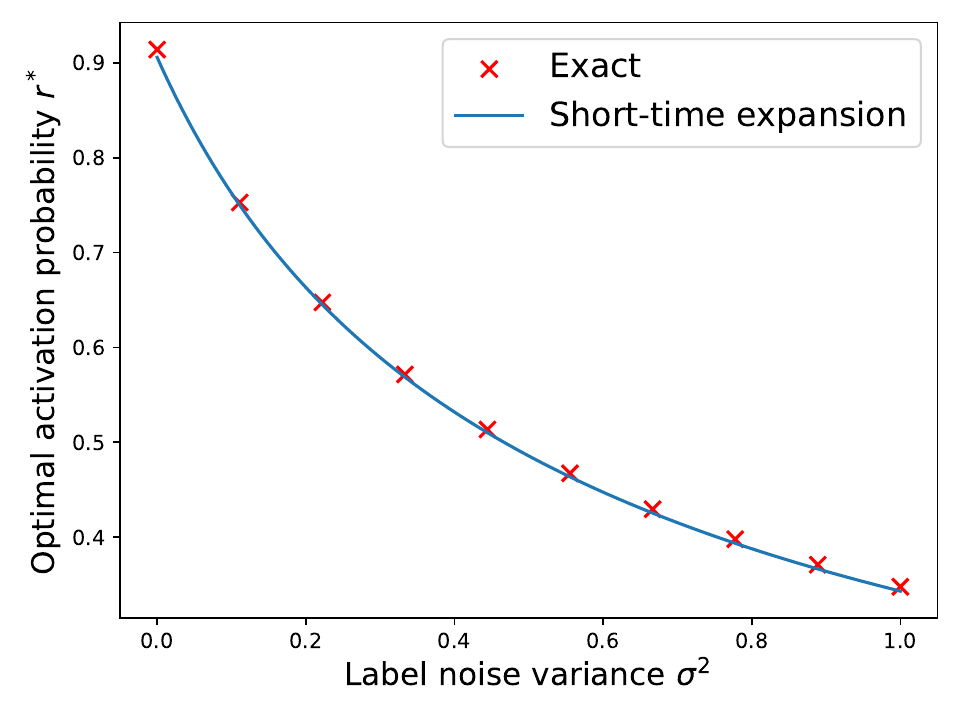}
  \caption{Optimal activation rate $r^*$ as a function of the variance $\sigma^2$ of the label noise. Parameters: $K=2$, $M=1$, $\eta=1$, $\eta_v=0$, $\alpha=0.1$. The teacher weight vector $w^*$ is taken to be of unit norm, the student weights $w$ are initialized to zero. The second layer weights are fixed to one. The continuous line represents the optimal rate obtained by numerical optimization by using the full equations of motion. The crosses correspond to the analytical result in Eq.~\eqref{eq:sort_time_drop}, obtained from a short-time expansion.}%figure generated with dropout-constant-optimal.ipynb
    \label{fig:optimal_const_dropout}
\end{figure}

First, in Fig.~\ref{fig:optimal_const_dropout} we examine the optimal dropout rate as a function of the label noise variance $\sigma^2$. Here, the crosses are obtained by numerical optimization of the full dynamical equations, while the continuous line corresponds to the asymptotic result derived analytically in Sec.~\ref{sec:early_times}. In the absence of label noise ($\sigma = 0$), the optimal activation rate $r^*$ remains close to unity, indicating minimal dropout. Conversely, as the label noise increases, the optimal strategy shifts toward a higher dropout probability, corresponding to a lower activation probability. 

As we will show in the following sections, the fact that the optimal activation rate $r^*$ decreases with increasing label noise holds robustly across the different phases of the learning dynamics.

\subsection{Early-time analysis}
\label{sec:early_times}

A first case in which the equations of motion can be solved exactly is in the early-time regime. In this section, we derive the optimal dropout rate in this limit. We assume that the student parameters are initialized with zero norm, i.e., $Q_{ii} = 0$, although the calculation can be readily extended to include a finite initial norm. At short times, the overlaps remain close to a symmetric configuration, where each student node is equally aligned with each of the teacher nodes. In other words,
\begin{equation}
    w_i = R \sum_n w^*_n + \tilde{w}_i\,, \label{eq:symm_ansatz_wi_perp}
\end{equation}
where, with a slight abuse of notation, $R$ denotes a positive scalar and $\tilde{w}_i$ is an additional node-dependent random vector, perpendicular to $w^*_n$ for all $n$. This motivates the ansatz
\begin{equation}
   Q_{ik} = Q\, \delta_{ik} + C\, (1 - \delta_{ik})\,, \qquad R_{in} = R\,, \label{eq:early_time_ansatz}
\end{equation}
where again $Q$ and $R$ denote positive scalar quantities.

At the initial time $\alpha=0$, the order parameters are $Q=C=R=0$ and the dynamical equations become, to leading order,
\begin{align}
    \frac{\mathrm{d} R}{\mathrm{d} \alpha}\approx\frac{\sqrt{2}}{\pi}\eta r\,,\quad\frac{\mathrm{d} Q}{\mathrm{d} \alpha}\approx\frac{2}{\pi}\eta^2 r\left(\frac{M}{3}+\sigma^2\right)\,,\quad\frac{\mathrm{d} C}{\mathrm{d} \alpha}\approx\frac{2}{\pi}\eta^2 r^2\left(\frac{M}{3}+\sigma^2\right)\,.
\end{align}
Therefore, at short times, the order parameters evolve independently as
\begin{equation}
R(\alpha)\approx\frac{\sqrt{2}}{\pi}\eta r\alpha,\quad Q(\alpha)\approx\frac{2}{\pi}\eta^2 r\left(\frac{M}{3}+\sigma^2\right)\alpha\,,\quad C(\alpha)\approx\frac{2}{\pi}\eta^2 r^2\left(\frac{M}{3}+\sigma^2\right)\alpha\,.
\end{equation}
Plugging these expressions into Eq.~\eqref{eq:gen_err} and expanding for small $\alpha$, we find
\begin{equation}
    \epsilon_g\approx \frac{2}{\pi^2}\eta^2 K (M/3+\sigma^2)((K-1)r^4+r^3)\alpha-\frac{2}{\pi^2}K M \eta r^2\alpha\,.\label{eq:short_time_gen_err}
\end{equation}
Minimizing the right-hand side with respect to the activation probability $r$, we find the unique positive solution
\begin{equation}
    r^*=\frac{12 M}{3M \eta+9\eta \sigma^2+\sqrt{3\eta (M+3\sigma^2)(M(32(K-1)+3\eta)+9\eta \sigma^2)}}\,.\label{eq:sort_time_drop}
\end{equation}
In Fig.~\ref{fig:optimal_const_dropout}, we compare this early-time result (continuous line) with the exact result obtained by numerically optimizing the full dynamics with $\alpha=0.1$ (crosses), finding good agreement.
As expected, the activation rate decreases with increasing label variance $\sigma^2$. For large $\sigma^2$ \rebuttal{, i.e., for $\sigma^2\gg M \max[1/3,32(K-1)/(9\eta)]$ }, we find $r^*\approx 2M/(3\eta \sigma^2)$.

\begin{figure}
    \centering
    \includegraphics[width=0.7\linewidth]{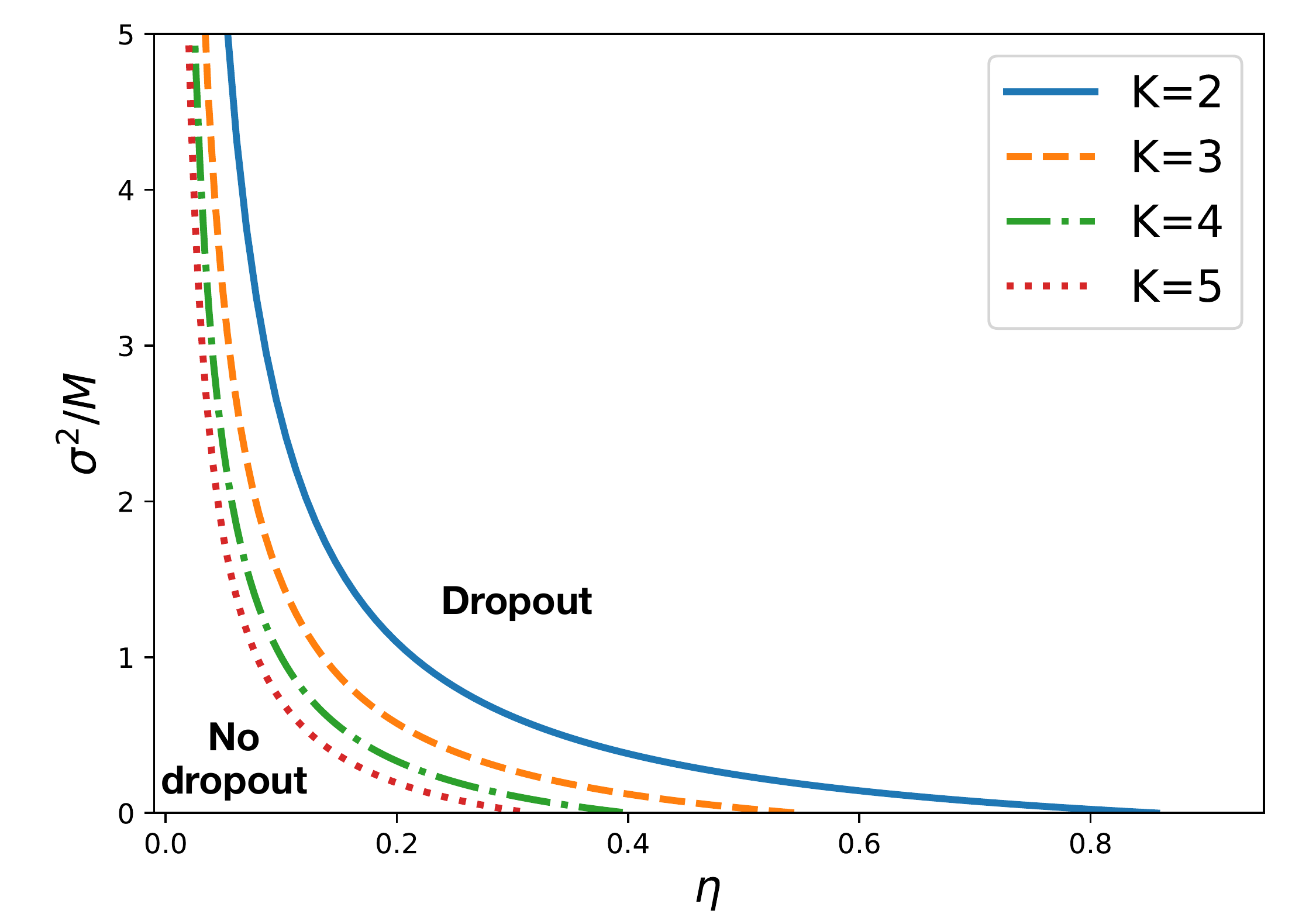}
  \caption{{\bf Early-times phase diagram.} Above the lines, obtained for different values of the number $K$ of hidden nodes in the student, it is beneficial to use dropout, with the optimal activation probability given in Eq.~\eqref{eq:sort_time_drop}. Below the line, it is optimal to avoid dropout, corresponding to activation probability $r=1$.}
    \label{fig:phase_digram}
\end{figure}

Imposing $r^*\leq 1$, we find the additional condition
\begin{equation}
    \sigma^2\geq M\left[ \frac{2}{(4K-1)\eta}-\frac13\right]\,.\label{eq:short_time_condition_dropout}
\end{equation}
Therefore, the optimal activation probability is $r^*=1$ (no dropout) below a threshold value of the label noise, and it is given by the expression in Eq.~\eqref{eq:sort_time_drop} otherwise. Note that the right-hand side of Eq.~\eqref{eq:short_time_condition_dropout} is positive only if
\begin{equation}
    \eta<\frac{6}{4K-1}\,.
\end{equation}
These observations lead to the phase diagram in Fig.~\ref{fig:phase_digram}. We observe that dropout is beneficial at high values of the learning rate or the label noise. Interestingly, the region where dropout is useful increases for increasing number of student's hidden nodes, in agreement with the intuition that dropout prevents negative co-adaptations.

Note that if we jointly optimize the generalization error in Eq.~\eqref{eq:short_time_gen_err} with respect to both the learning rate and the activation probability, we find $r^*=1$ and $\eta=M/(2K(\sigma^2+1/3)$. This implies that, at short times, dropout is not necessary provided that the learning rate is properly tuned.

\subsection{Intermediate-time analysis: unspecialized phase}\label{sec:uspecialized}

By numerically integrating the ODEs~\eqref{eq:ODE_dropout_main}, we observe that the generalization error exhibits a plateau at intermediate times. This plateau corresponds to the \rebuttal{\emph{unspecialized}---or \emph{symmetric}---phase}, previously described in Refs.~\cite{biehl1995learning,saad1995PRL,saad1995PRE,goldt2019dynamics,goldt2020hmm}. \rebuttal{This phase is characterized by a symmetry of} the student weight vectors $w_i$\rebuttal{, that} are equally aligned with all teacher weights. This is in contrast to the \emph{specialized phase} that emerges at later times, where each student node $w_i$ specializes by aligning predominantly with a specific teacher weight $w^*_n$ or becoming inactive. Note that if the second layer is also trained, more complex specialization patterns can emerge, as the network gains additional flexibility in how it distributes representational roles across the hidden units.

In this section, we analytically investigate the role of dropout during the intermediate phase of training. Despite the complexity of the ODEs, exact results can be obtained in the limit of small learning rate $\eta \to 0$, allowing us to derive expressions for both the optimal dropout rate and the corresponding generalization error. We then compute perturbative corrections around this limit to account for finite learning rates. Finally, we examine how dropout influences the transition to specialization. As we will show, this transition only takes place when the activation probability $r$ exceeds a critical threshold $r_c$.

\subsubsection{Small-$\eta$ regime}
\label{sec:fixed_point_low_order}

\begin{figure}[t!]
    \centering
    \includegraphics[width=0.45\linewidth]{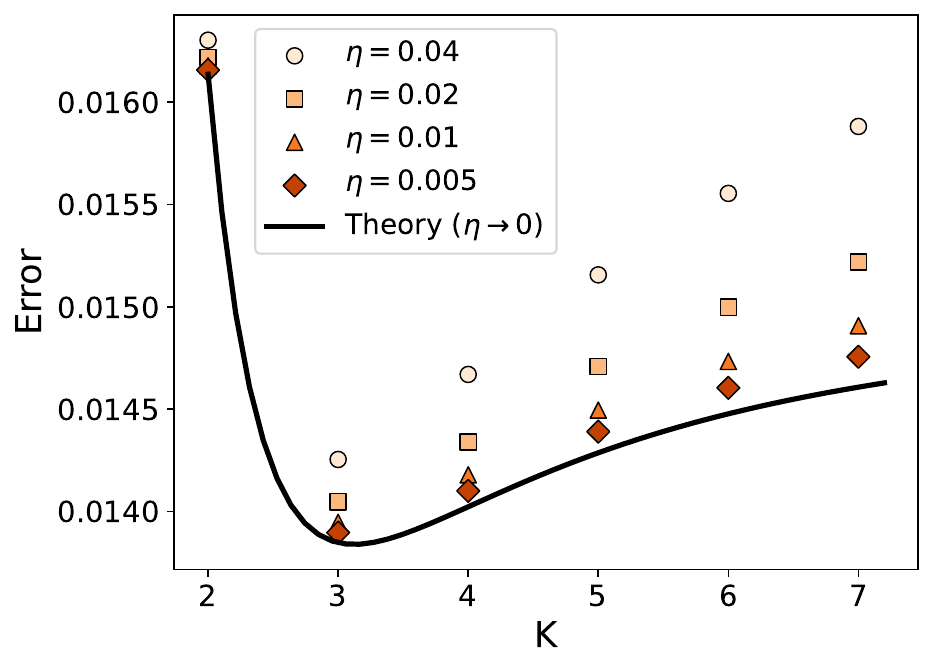}
     \includegraphics[width=0.45\linewidth]{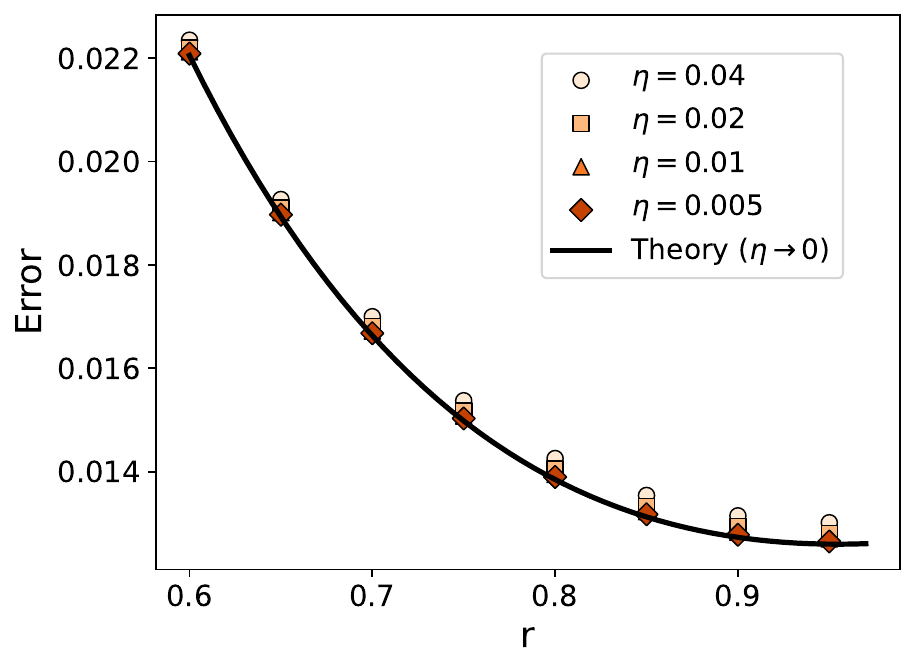}
    \caption{\textbf{Left.} Generalization error at the plateau vs. the student's width $K$ at fixed activation rate $r=0.8$. \textbf{Right.} Generalization error at the plateau vs. the activation rate $r$ at fixed student's width $K=3$. In both panels, the symbols are obtained by numerical integration of Eq.~\eqref{eq:ODE_dropout_main} at different values of the learning rate $\eta$ and with $\sigma=0$. The black line marks the theoretical prediction in the small-$\eta$ regime from Eq.~\eqref{eq:epsg_plateau}. The teacher's width is fixed to $M=2$.}
    \label{fig:low_eta_theory}
\end{figure}

In this section, we investigate how dropout affects the performance of the network in the unspecialized phase. To do so, we identify fixed-point solutions of the ODEs~\eqref{eq:ODE_R_dropout}--\eqref{eq:ODE_Q_dropout}. To make analytic progress, we first consider the small-$\eta$ limit and neglect terms of order $\eta^2$ in the ODEs. Notably, this approximation eliminates the dependence on the label noise variance $\sigma^2$. We will later perform a perturbative expansion in $\eta$ to recover this dependence to leading order. By setting $f_{R_{in}}=0$ and $f_{Q_{ik}}=0$ to leading order in $\eta$, we find Eqs.~\eqref{eq:leadingR} and \eqref{eq:leadingQ}. Since we are interested in an unspecialized solution, we assume the student weights take the form
\begin{equation}
    w_i = R_0 \sum_n w^*_n\,, \label{eq:symm_ansatz_wi}
\end{equation}
for some scalar $R_0$ to be determined. In other words, for small learning rate $\eta$, we expect the student weight vectors to lie entirely within the subspace spanned by the teacher vectors $\{w^*_n\}$, without acquiring an additional component in the orthogonal direction, as in Eq.~\eqref{eq:symm_ansatz_wi_perp}. \rebuttal{In other words, in the small-$\eta$ limit the student does not overfit the noise. This can be verified by checking that the corresponding ansatz}
\begin{equation}
    Q_{ik} = R_0^2 M\,, \qquad R_{in} = R_0\,
\end{equation}
\rebuttal{is a fixed point of the ODEs.} By inserting this ansatz into Eqs.~\eqref{eq:leadingR} and~\eqref{eq:leadingQ}, we find that it indeed satisfies the fixed-point equations, yielding
\begin{equation}
  \hspace{-.8em} \frac 1R_0= {  \sqrt{ \left( 1 + (K-1)r \right)^2 -M + \sqrt{ M^2 + (K-1)r \left(1 + (K-1)r\right)^2 \left(2 + (K-1)r\right) } } }\,.\label{eq:gamma0}
\end{equation}
Correspondingly, the generalization error, i.e., the height of the plateau, reads
\begin{equation}
    \epsilon_g^0 = \frac{1}{\pi} \left( \frac{M \pi}{6} + 
    K^2 r^2 \arcsin\left( \frac{M R_0^2}{1 + M R_0^2} \right) - 
    2 K M r \arcsin\left( \frac{R_0}{\sqrt{2} \sqrt{1 + M R_0^2}} \right) \right)\,.\label{eq:epsg_plateau}
\end{equation}
This expression is shown in Fig.~\ref{fig:low_eta_theory} as a function of $K$ and $r$ and is in good agreement with the results of numerically integrating the full ODEs \eqref{eq:ODE_dropout_main} at small but finite $\eta$. We observe that at fixed $r$ and $M$, the generalization error at the plateau shows a minimum at a finite value of $K$. Similarly, at fixed $M=2$ and $K=3$, the error is minimized at an intermediate value of the dropout rate, in this case $r^*\approx0.95$.

In the special case $M = K$ and $r = 1$, we recover the result of Ref.~\cite{saad1995PRE}
\begin{equation}
    \epsilon_g^0=\frac{K}{\pi}\left[\frac{\pi}{6}-K \operatorname{arcsin}\left(\frac{1}{2K}\right)\right]\,,
\end{equation}
which is here generalized to arbitrary $M$, $K$, and $r$. It is also interesting to consider the overparametrized limit $K\gg 1$ at fixed $M$ and $r$, leading to 
\begin{equation}
    \epsilon_g^0 \approx  \frac{\pi-3}{6\pi}M +\frac{11+12r(r-2)}{24\pi r^2}\frac{M}{K^2}+\mathcal{O}(\frac{1}{K^3})\,,
\end{equation}
which is minimized by the optimal activation probability $r^*_0=11/12\approx 0.92$, independently of $K$ and $M$.

\begin{figure}
    \centering
    \includegraphics[width=0.45\linewidth]{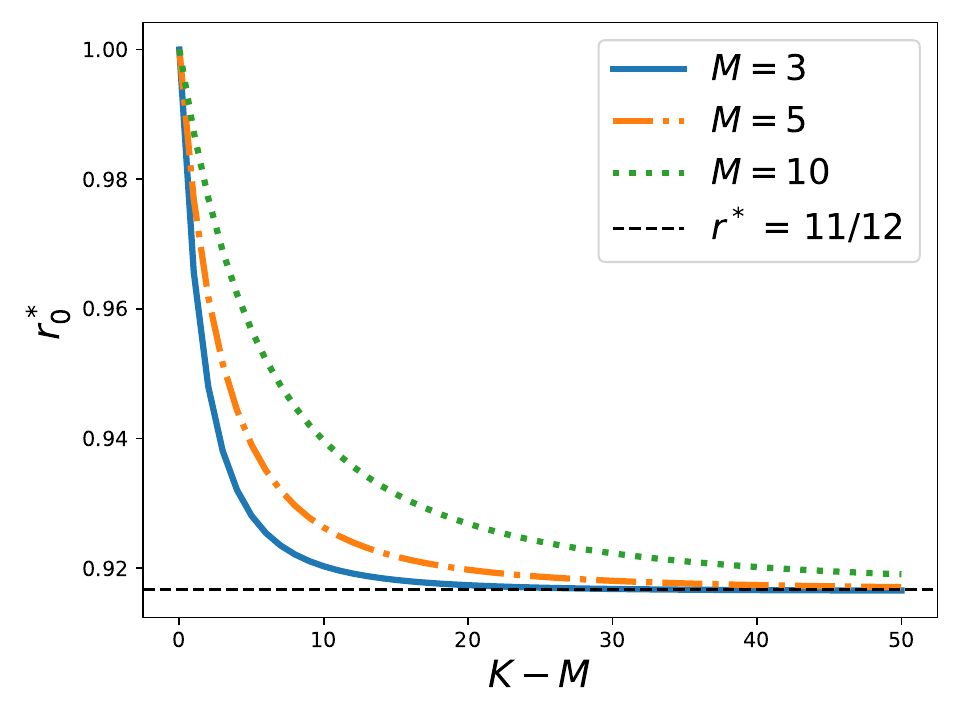}
    \includegraphics[width=0.45\linewidth]{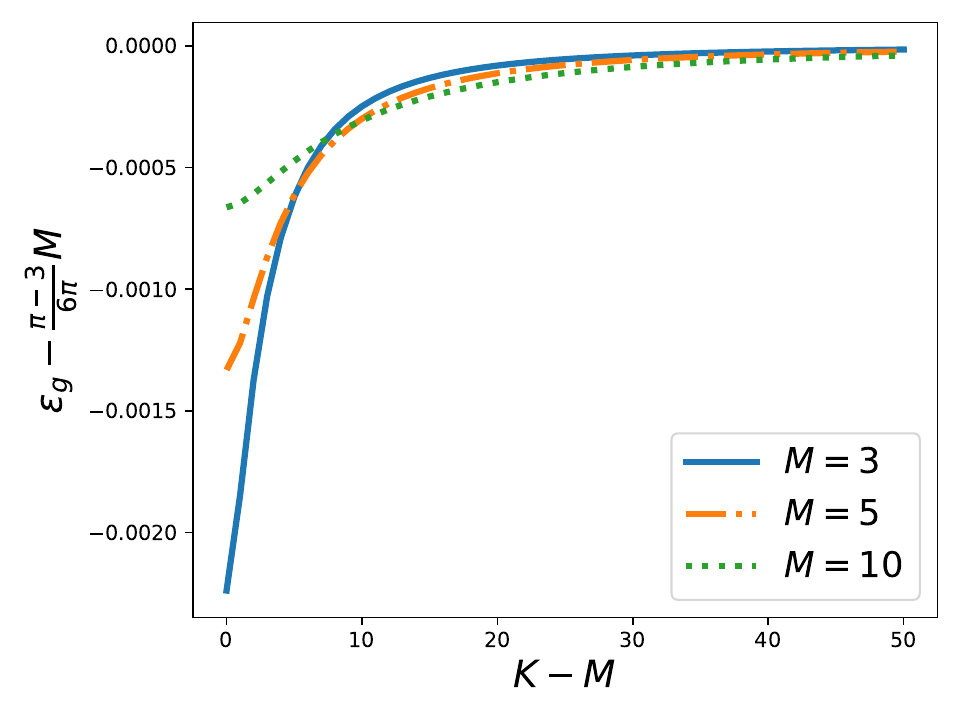}
  \caption{{\bf Left:} Optimal activation probability $r^*_0$ in the plateau phase as a function of $K - M$ for different values of $M$. The curves are obtained via numerical minimization of the generalization error in Eq.~\eqref{eq:epsg_plateau}, derived in the small-$\eta$ limit. For large $K$, all curves converge to the universal value $r^*_0 = 11/12$. {\bf Right:} Corresponding value of the shifted generalization error, $\epsilon_g - \frac{(\pi - 3)M}{6\pi}$, as a function of $K - M$. }
    \label{fig:rstar_plateu}
\end{figure}

The optimal activation probability $r^*_0$, obtained by minimizing Eq.~\eqref{eq:epsg_plateau}, is shown in Fig.~\ref{fig:rstar_plateu}. Interestingly, we observe that $r^*_0$ decreases with increasing $K$ and increases with $M$. This behavior can be intuitively understood: when the student has more hidden units (larger $K$), fewer of them need to be active at any given time to capture the relevant features; conversely, a more expressive teacher (larger $M$) demands a higher number of active units in the student to effectively represent the target function.

It is worth noting that the ansatz in Eq.~\eqref{eq:symm_ansatz_wi} constrains the student to lie in the $M$-dimensional subspace spanned by the teacher vectors $w^*_n$, for $n = 1, \ldots, M$. As a result, the student’s hidden units do not overfit the label noise, which is consistent with the behavior expected in the small-$\eta$ regime. For finite $\eta$, however, each vector $w_i$ acquires an additional component $\tilde{w}_i$—orthogonal to the teacher subspace—as described by Eq.~\eqref{eq:symm_ansatz_wi_perp}, leading to noise-fitting.

Although the small-$\eta$ analysis does not capture this noise-fitting component, we find, somewhat surprisingly, that dropout can still improve performance in this regime, as shown in Fig.~\ref{fig:rstar_plateu}. However, the optimal activation probability remains close to one ($r^* > 11/12$), and the gain in generalization error is modest. We expect the effect of dropout to be more pronounced in the finite-$\eta$ regime, where noise-fitting becomes significant and can degrade performance. This observation motivates the perturbative expansion presented in the next section, aimed at computing leading-order corrections in $\eta$ to the results discussed here.

As a final remark, we note that while choosing a small value of $\eta$ mitigates noise-fitting, it also slows down convergence, resulting in a tradeoff between training speed and precision that we do not explore in detail in this work.

\subsubsection{Perturbative corrections to the small-$\eta$ regime}
\label{sec:perturbative}

In this section, we compute the first-order correction to the small-$\eta$ results. As discussed above, for finite $\eta$ the student weight vectors are expected to acquire additional random components due to noise fitting. This motivates the refined ansatz
\begin{equation}
    w_i = R \sum_{n=1}^M w^*_n + \tilde{w}_i\,,
\end{equation}
where $\tilde{w}_i$ is a random vector orthogonal to the teacher subspace and $R$ is a scalar. The effect of dropout in this regime is to reduce correlations among the noise components $\tilde{w}_i$ for $i = 1, \ldots, K$, promoting error cancellation and thereby improving generalization.

The corresponding ansatz for the order parameters is
\begin{equation}
    Q_{ik} = Q\, \delta_{ik} + C\, (1 - \delta_{ik})\,, \qquad R_{in} = R\,.
\end{equation}
To carry out the perturbative expansion, we expand the order parameters to linear order in $\eta$:
\begin{equation}
    Q = Q_0 + \eta Q_1\,, \qquad C = Q_0 + \eta C_1\,, \qquad R = R_0 + \eta R_1\,,
\end{equation}
with $Q_0 = R_0^2 M$ and $R_0$ is defined in Eq.~\eqref{eq:gamma0}.

Using \textit{Mathematica}, we insert this ansatz into the full ODEs given in Eqs.~\eqref{eq:ODE_R_dropout}--\eqref{eq:ODE_Q_dropout} and expand to leading order in $\eta$. This yields analytical expressions for the first-order corrections $Q_1$, $C_1$, and $R_1$. These expressions are rather lengthy and not particularly illuminating, so we do not report them here. However, we provide a detailed \textit{Mathematica} notebook at \cite{dropout_notebooks} that reproduces the derivation and enables the user to compute the explicit forms of the corrections for arbitrary values of $K$, $M$, and $r$.

\begin{figure}
    \centering
    \includegraphics[width=0.45\linewidth]{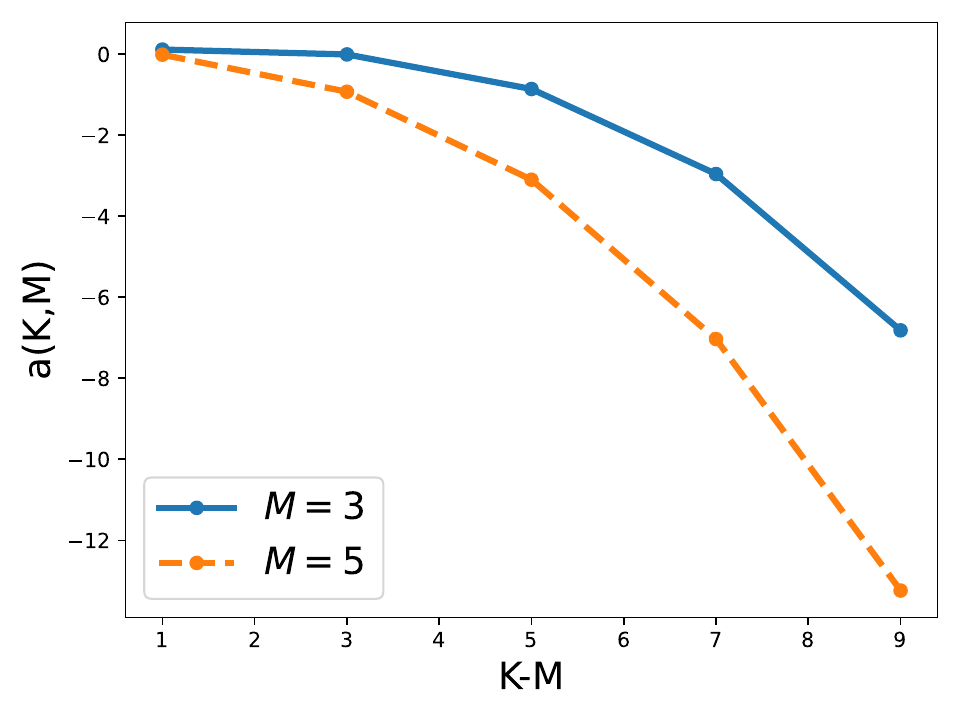}
    \includegraphics[width=0.45\linewidth]{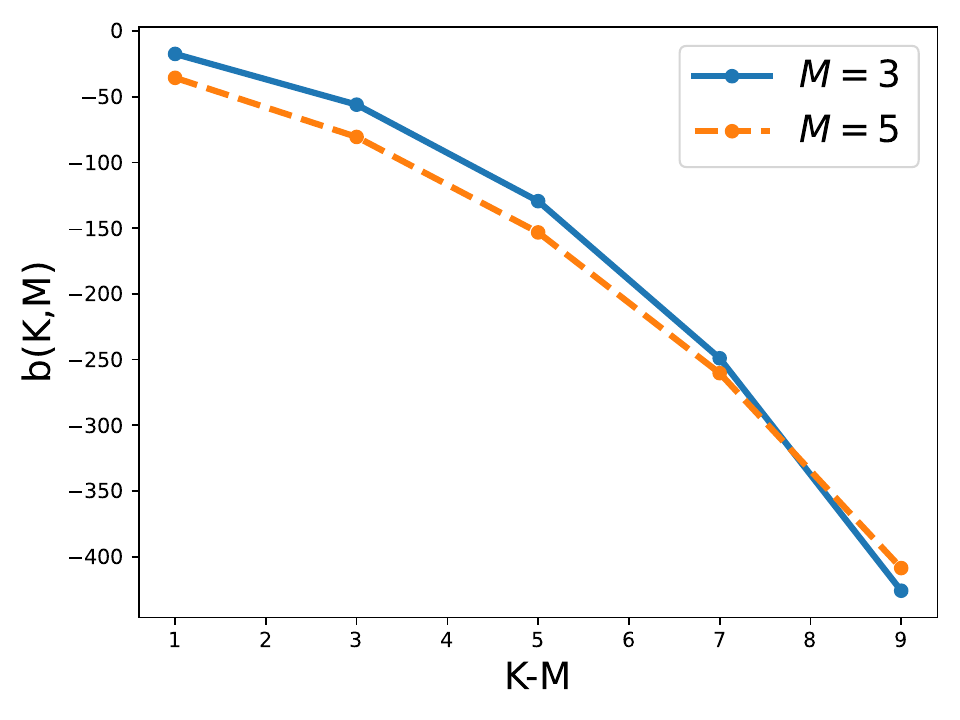}

\caption{Coefficients $a(K,M)$ (left) and $b(K,M)$ (right) \rebuttal{for the plateau phase} as a function of $K-M$ for different values of $M$.  }
    \label{fig:coeff}
\end{figure}

To leading order in $\eta$, the generalization error takes the form
\begin{equation}
    \epsilon_g \approx \epsilon_g^0(r) + \eta\, \epsilon_g^1(r, \sigma)\,,
\end{equation}
where $\epsilon_g^0(r)$ is given in Eq.~\eqref{eq:epsg_plateau}, and an explicit expression for $\epsilon_g^1(r, \sigma)$ is provided in the notebook at \cite{dropout_notebooks}. Recalling that $r_0^*$ denotes the minimizer of $\epsilon_g^0(r)$, we find that the optimal activation probability for small $\eta$ can be written as
\begin{equation}
    r^* \approx r_0^* + r_1^*(\sigma)\, \eta\,,
\end{equation}
where the linear correction is given by
\begin{equation}
    r_1^*(\sigma) = -\frac{\partial_r \epsilon_g^1(r_0^*, \sigma)}{\partial_r^2 \epsilon_g^0(r_0^*)} = a(K, M) + b(K, M)\, \sigma^2\,,
\end{equation}
where the coefficients $a(K, M)$ and $b(K, M)$ are available at \cite{dropout_notebooks}. For instance, in the case $K = 2$ and $M = 1$, we find $r_0^* = 0.931\ldots$, $a(2,1) = 0.023\ldots$, and $b(2,1) = -6.445\ldots$. In Fig.~\ref{fig:coeff}, we plot the coefficients $a(K, M)$ and $b(K, M)$ for various values of $M$ and $K$. We observe that both coefficients are negative for most values of $K$ and $M$, although $a(K, M)$ becomes positive when both $K$ and $M$ are small. This suggests that, in general, increasing the label noise variance $\sigma^2$ leads to a decrease in the optimal activation probability $r$.

\subsection{Onset of specialization}
\label{sec:linear_stability}

In this section, we investigate how the specialization transition is affected by the presence of dropout. Recall that in the absence of dropout ($r = 1$) this transition has been characterized in Refs.~\cite{saad1995PRE,biehl1996transient,straat2019line}, where it was shown that the symmetric, unspecialized phase is linearly unstable. Small perturbations—arising either from noise or from the initial conditions—grow over time, ultimately driving the system toward a specialized phase in which each student node aligns predominantly with a distinct teacher node or is deactivated. For example, in the case $K > M$, a typical specialized configuration takes the form $w_i \approx w^*_i$ for $i \leq M$ and $w_i = 0$ for $i > M$. The precise dependence of this specialization transition on the initialization of the online dynamics has been investigated in \cite{jarvis2025a}. \rebuttal{Moreover, changing the scaling of the student readout can affect the rate of convergence toward specialization \cite{PhysRevE.105.L052302}.} Note that the hidden-unit specialization transition has also been studied within the asymptotic Bayes-optimal framework, as discussed, for example, in Refs.~\cite{oostwal2021hidden,aubin2018committee}.

Here, we analyze how the introduction of dropout modifies this transition to specialization. For simplicity, we focus on the case $K=M$ and the small-$\eta$ regime. To proceed, we consider the ansatz
\begin{equation}\label{eq:ansatz_specialization}
    Q_{ij}=Q\delta_{ij}+C(1-\delta_{ij})\,,\quad R_{in}=R\delta_{in}+S(1-\delta_{in})\,,
\end{equation}
where the structure of the $R_{in}$ matrix now allows for specialization. Using this ansatz, the evolution equations for the order parameters then can be written in a simplified form, given in Eqs.~\eqref{eq:special_eqR}-\eqref{eq:special_eqC}.

As shown in Sec.~\ref{sec:fixed_point_low_order}, the plateau phase corresponds to the fixed point
\begin{equation}
    Q = C = R_0^2 M\,, \qquad R = S = R_0\,,
\end{equation}
where $R_0$ is given in Eq.~\eqref{eq:gamma0}. To study the stability of this fixed point, we perform a linear stability analysis by introducing small perturbations:
\begin{equation}
    Q = R_0^2 M + Q_1\,, \qquad C = R_0^2 M + C_1\,, \qquad R = R_0 + R_1\,, \qquad S = R_0 + S_1\,,
\end{equation}
where $Q_1$, $C_1$, $R_1$, and $S_1$ denote small deviations from the fixed-point values.

\begin{figure}
    \centering
    \includegraphics[width=0.7\linewidth]{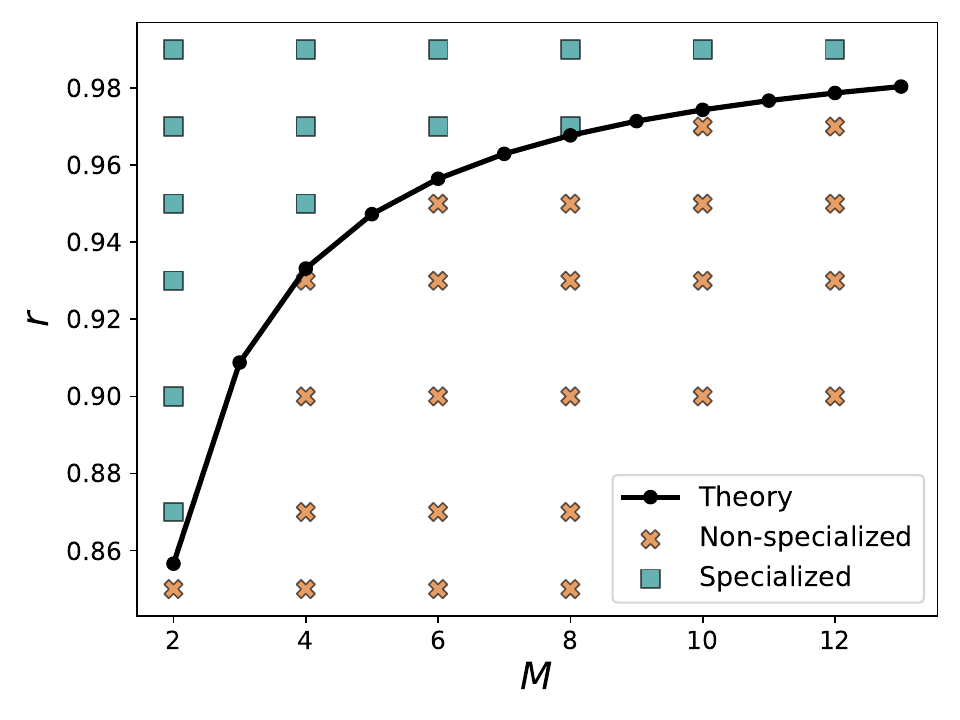}
  \caption{\textbf{Specialization phase diagram in the $(M, r)$ plane}. In the non-specialized phase, the plateau fixed point is linearly stable. As $r$ increases, the fixed point becomes unstable, signaling a transition to the specialized phase where student units align with individual teacher units. The continuous black line is the theoretical transition line obtained from linear stability analysis in the $\eta\to 0$ limit. The symbols are obtained by numerically integration of the full ODEs \eqref{eq:ODE_dropout_main} up to time $\alpha=50000$, with $\eta=0.01$, and $\sigma=0$. }
    \label{fig:phase_digram_special}
\end{figure}
Using \emph{Mathematica}, we expand the ODEs to linear order in the perturbations and obtain the following system of equations:
\begin{equation}
    \frac{\mathrm{d}}{\mathrm{d}\alpha}
    \begin{pmatrix}
        R_1 \\
        S_1 \\
        Q_1 \\
        C_1
    \end{pmatrix}
    = A(M, r)
    \begin{pmatrix}
        R_1 \\
        S_1 \\
        Q_1 \\
        C_1
    \end{pmatrix}\,,
\end{equation}
where $A(M, r) \in \mathbb{R}^{4 \times 4}$ is a matrix whose entries depend on $M$ and $r$. The explicit form of $A(M, r)$ is lengthy and provided in the accompanying \emph{Mathematica} notebook at \cite{dropout_notebooks}.

By inspecting the eigenvalues of $A(M,r)$, we find the phase diagram in Fig.~\ref{fig:phase_digram_special}, where we identify two phases: a non-specialized phase where the plateau fixed point is linearly stable, and a specialized phase which is reached if the fixed point becomes unstable. The theoretical line corresponds to the critical activation probability $r_c$ at which the real part of the largest eigenvalue of $A(M, r)$ vanishes. It shows excellent agreement with the results obtained by numerically integrating the full ODEs in Eq.~\eqref{eq:ODE_dropout_main}. Let us also stress that in the specialized phase the student vectors will not perfectly align with the teacher weights, however they will show a preferential overlap with one of the teacher weights.

\subsection{Long-time behavior}\label{sec:long_time}

\subsubsection{\rebuttal{Perturbative expansion for small $\eta$ ($K=M$)}}

\begin{figure}
    \centering
    \includegraphics[width=0.45\linewidth]{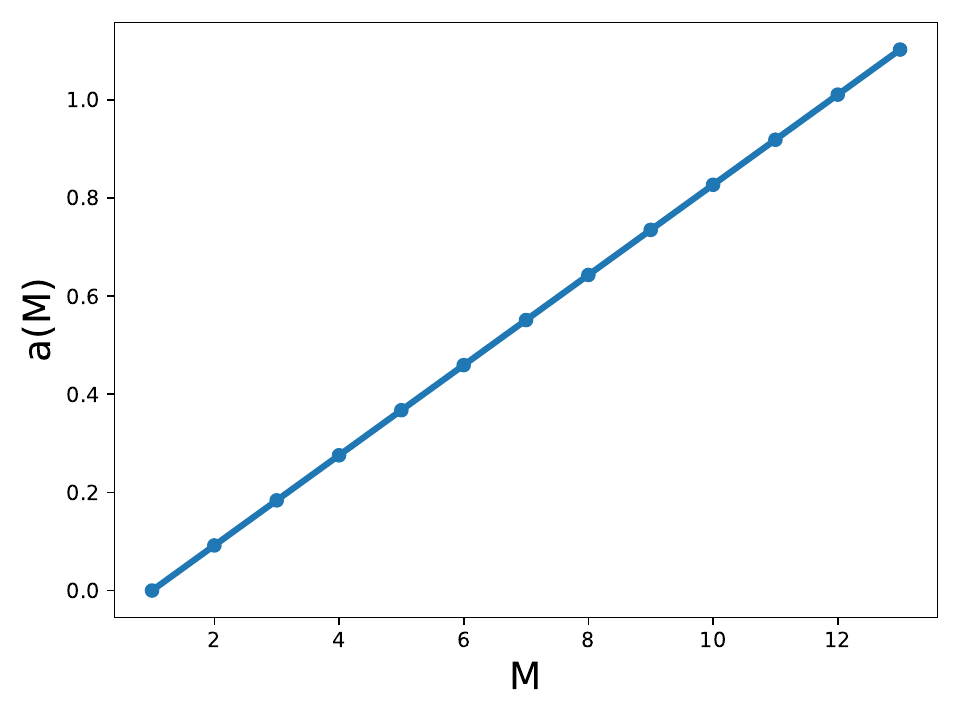}
    \includegraphics[width=0.45\linewidth]{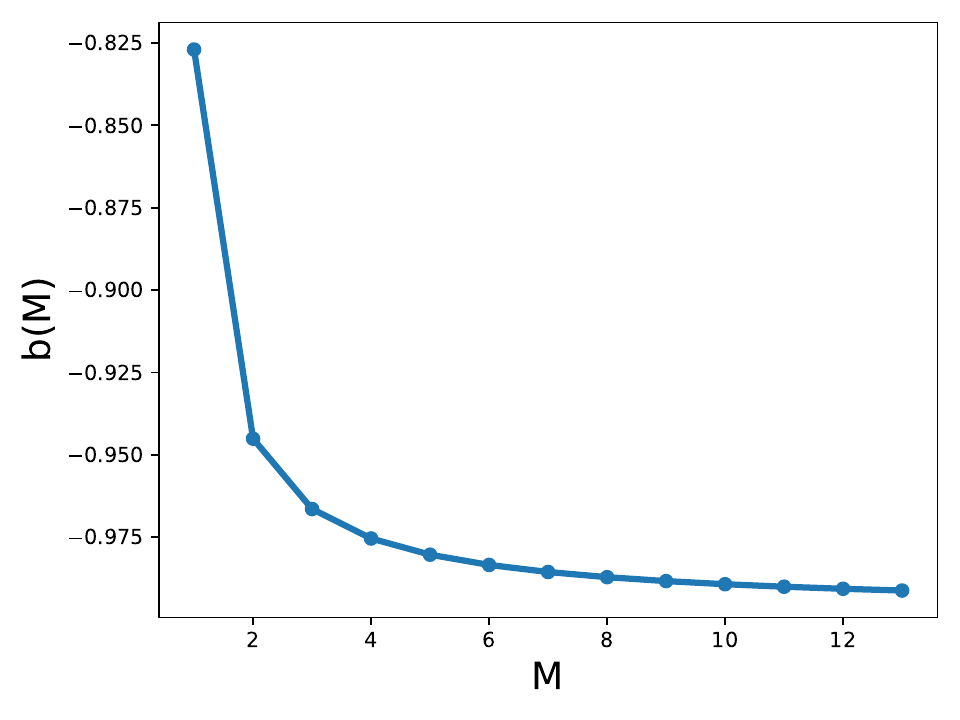}

\caption{\rebuttal{Coefficients $a(M)$ (left) and $b(M)$ (right) as a function of $M$ for the specialized phase with $K=M$. } }
    \label{fig:coeff_specialized}
\end{figure}

\rebuttal{To investigate the long-time behavior, we first focus on the case $K=M$ and the small-$\eta$ regime. In the limit $\eta\to0$, perfect specialization, corresponding to a student perfectly reproducing the teacher with vanishing generalization error, is obtained if $r=1$. In this limit, the specialized order parameters are $Q_{ij}=R_{ij}=\delta_{ij}$. }

\rebuttal{We compute the first-order corrections in $\eta$. The ansatz in this case is
\begin{align}
    Q_{ij}=\delta_{ij}+\eta \delta_{ij} Q_1 +\eta (1-\delta_{ij}) C_1\,\quad R_{ij}=\delta_{ij}+\eta \delta_{ij} R_1 +\eta (1-\delta_{ij}) S_1\,.
\end{align}
One can then perform a perturbative calculation identical to that in Section \ref{sec:perturbative}. We find that in the small-$\eta$ limit the optimal activation probability is given by
\begin{equation}
    r^*\approx\min\left[1; \, 1+\eta(a(M)+\sigma^2 b(M))\right]\,,
\end{equation}
The coefficients $a(M)$ and $b(M)$ are shown in Fig.~\ref{fig:coeff_specialized} and can be computed for arbitrary $M$ via the corresponding \emph{Mathematica} notebook at \cite{dropout_notebooks}. Once again, we observe that the coefficient $b(M)$ is negative, implying that larger noise $\sigma$ requires setting lower values of the activation probability.}

\subsubsection{\rebuttal{Numerical analysis}}

\begin{figure}
    \centering
\includegraphics[width=0.32\linewidth]{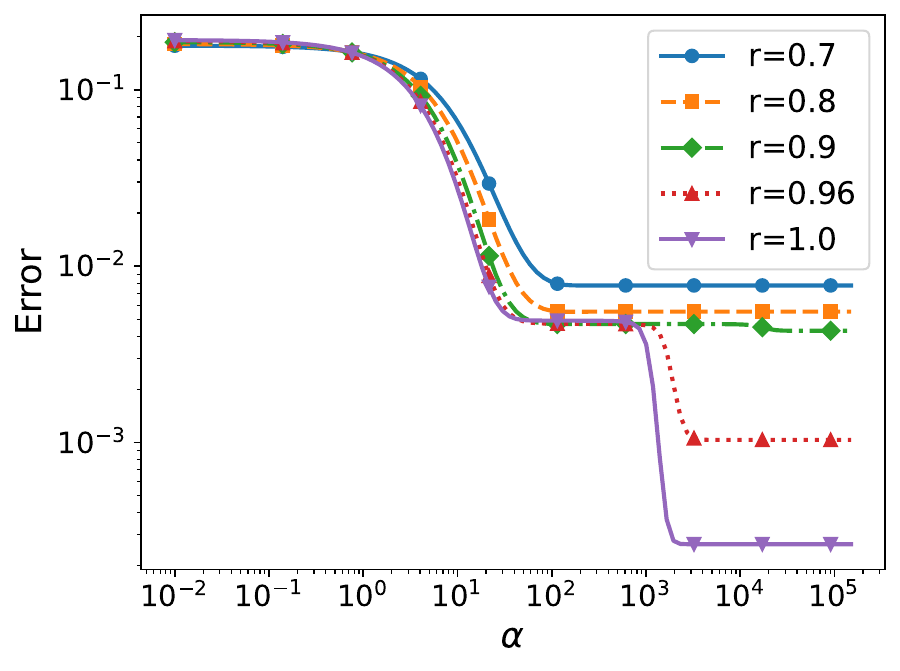}
\includegraphics[width=0.32\linewidth]{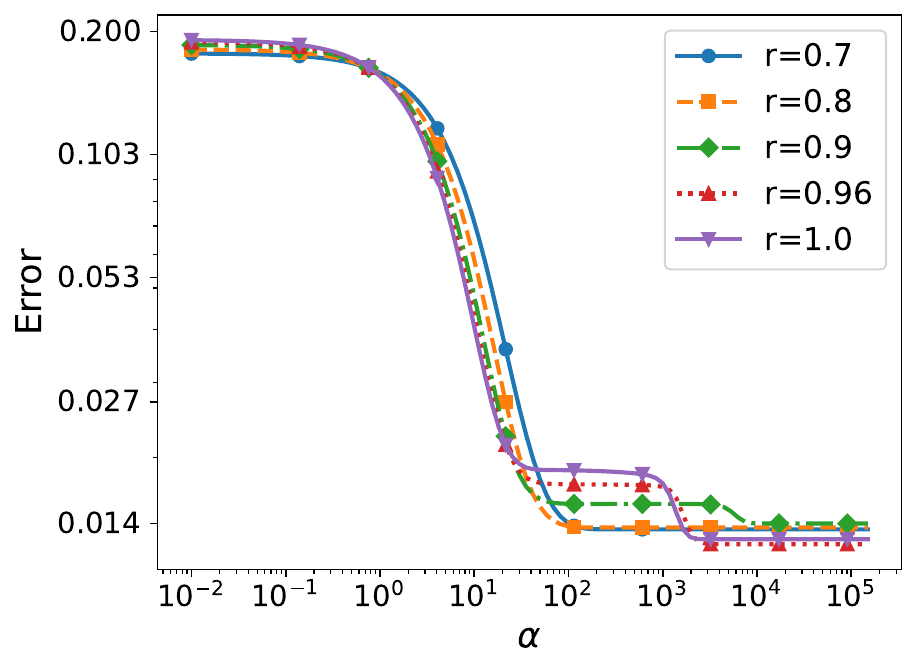}
\includegraphics[width=0.32\linewidth]{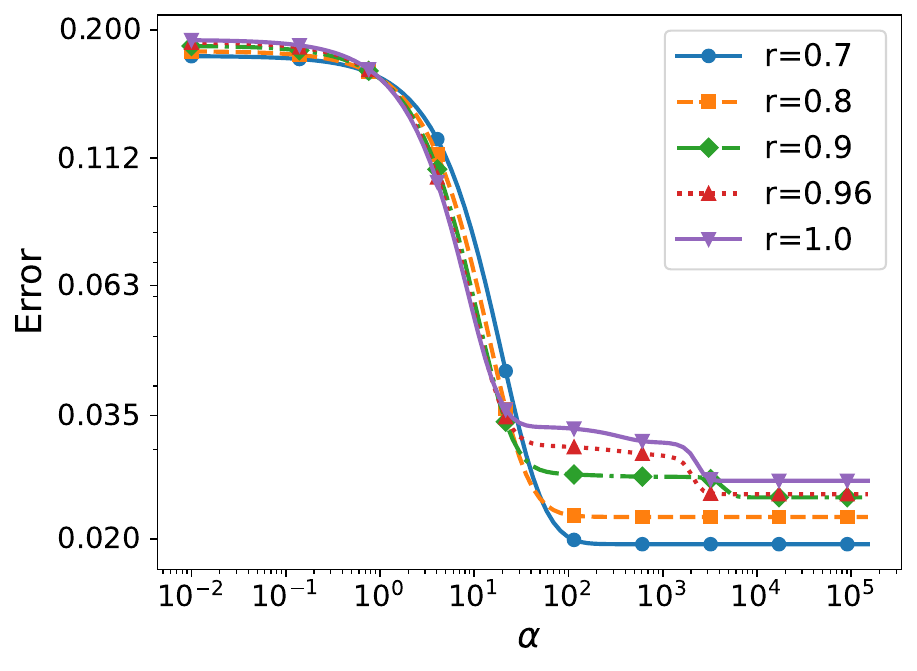}
    \caption{Generalization error as a function of time $\alpha$ for $K=2$, $M=1$, $\eta=0.1$, and different values of node-activation rate $r=0.7,0.8,0.9,0.96,1$. Each panel represents a different value of label noise $\sigma=0.1$ (left), $\sigma=0.7$ (middle), $\sigma=1$ (right). }
    \label{fig:error_plateau}
\end{figure}

As shown in section \rebuttal{\ref{sec:linear_stability}}, the long-time behavior depends strongly on the activation probability $r$. For $r < r_c$, the plateau phase remains stable and the network stays trapped in this phase even at long times. Conversely, specialization occurs at long times when $r > r_c$. As a result, the optimal activation probability $r^*$ depends on both the total training time $\alpha$ and the label noise variance $\sigma^2$. This behavior is summarized in Fig.~\ref{fig:error_plateau}.

For small label noise (left panel), a strategy with little or no dropout ($r \approx 1$) is optimal both before and after the specialization transition. In the opposite case of large noise (right panel), even though a large value of $r$ (e.g., $r = 0.96$) induces specialization, the resulting error is larger than that of an unspecialized configuration achieved with a smaller $r$ (e.g., $r = 0.7$), reflecting the beneficial effects of dropout at higher noise levels.

Interestingly, at intermediate noise levels (central panel), the optimal strategy depends on the total training time $\alpha$. At intermediate times (before the specialization transition), a lower value of $r$ is preferable. However, for larger $\alpha$, it is preferable to choose a higher $r$ to allow specialization and achieve a lower final error.

\section{\label{sec:discussion}Conclusions}

In this paper, we have studied the effects of dropout regularization on the training dynamics of two-layer neural networks in the teacher-student setting. We have derived exact closed-form equations that describe the training dynamics under online SGD in the high-dimensional limit. By analyzing these equations, we have shown that dropout helps reduce detrimental correlations between the nodes of the network. Furthermore, we have studied the optimal dropout rate and the corresponding generalization error, deriving several analytical results at short, intermediate, and long times. Our findings demonstrate that dropout can significantly mitigate the negative effects of label noise, and that the optimal activation probability decreases as the label noise increases. Finally, we have shown analytically that high dropout rates prevent the network from reaching a specialized configuration.

Our results rely on the assumption of online training, where a new sample is used at each iteration. An interesting direction for future work is to consider offline training, where the same samples are revisited multiple times. This would enable the study of how dropout affects memorization and overfitting. This analysis could be carried out using dynamical mean-field theory~\cite{agoritsas2018out,mignacco2020dynamical}.

While here we have focused on constant dropout strategies, an interesting open question is how to find optimal dropout schedules that vary the dropout probability over the course of training. Such schedules have been investigated empirically in Refs.~\cite{morerio2017curriculum,liu2023dropout}, where they were shown to improve generalization performance across various settings. Identifying optimal dropout schedules could be possible using the optimal control framework introduced in Ref.~\cite{mori2024optimal}. Such adaptive strategies will be analyzed in a separate publication~\cite{in_preparation}.

\subsection*{Acknowledgments}

This work was supported by a Leverhulme Trust International Professorship grant (Award Number: LIP-2020-014) and by the Simons Foundation (Award Number: 1141576).

\bibliography{biblio}
\bibliographystyle{unsrt}
\appendix

\section{Equations for the training dynamics}\label{appendix:derivationODE}

In this appendix, we present the dynamical equations discussed in the main text. The derivation of these equations follows closely that of Ref.~\cite{goldt2019dynamics}, with the difference that an additional binary variable is present to describe dropout. To this end, it is useful to define the notation
\begin{equation}
    \mathcal{N}\left[r,\{i,j,k,\ldots ,l\}\right] = r^n\,,
\end{equation}
where \( n = |\{i,j,k,\ldots ,l\}| \) denotes the cardinality of the set \(\{i,j,k,\ldots ,l\}\), i.e., the number of distinct indices in the set. Following \cite{goldt2019dynamics}, we show that, for $N,\mu\to\infty$ with $\alpha=\mu/N$ fixed,
\begin{align}
\frac{\mathrm{d} R_{in}}{\mathrm{d} \alpha} &= f_{R_{in}}\equiv \eta v_i \left[ 
    \sum_{m=1}^M \mathcal{N}\left[r,\{i\}\right]v_m^* I_3(i, n, m) - \sum_{j=1}^K  \mathcal{N}\left[r,\{i,j\}\right] v_j I_3(i, n, j)
\right], \label{eq:ODE_R_dropout} \\
\frac{\mathrm{d} Q_{ik}}{\mathrm{d} \alpha} &= f_{Q_{ik}}\equiv \eta v_i  \left[ 
    \sum_{m=1}^M \mathcal{N}\left[r,\{i\}\right] v_m^* I_3(i, k, m) - \sum_{j=1}^K \mathcal{N}\left[r,\{i,j\}\right] v_j I_3(i, k, j)
\right] \notag \\
&\quad + \eta v_k \left[ 
    \sum_{m=1}^M \mathcal{N}\left[r,\{k\}\right] v_m^* I_3(k, i, m) - \sum_{j=1}^K \mathcal{N}\left[r,\{k,j\}\right] v_j I_3(k, i, j)
\right] \notag \\
&\quad + \eta^2 v_i v_k  \Bigg[ 
    \sum_{n=1}^M \sum_{m=1}^M \mathcal{N}\left[r,\{i,k\}\right] v_n^* v_m^* I_4(i, k, n, m)  \notag \\
    & \quad\quad - 
    2 \sum_{j=1}^K \sum_{n=1}^M \mathcal{N}\left[r,\{i,k,j\}\right] v_j v_n^* I_4(i, k, j, n) \notag\\
&\quad\quad + \sum_{j=1}^K \sum_{l=1}^K \mathcal{N}\left[r,\{i,j,k,l\}\right] v_j v_l I_4(i, k, j, l) + \mathcal{N}\left[r,\{i,k\}\right] \sigma^2 J_2(i, k)
\Bigg], \label{eq:ODE_Q_dropout} \\
\frac{\mathrm{d} v_i}{\mathrm{d} \alpha} &= \eta_v \left[ 
    \sum_{n=1}^M \mathcal{N}\left[r,\{i\}\right] v_n^* I_2(i, n) - \sum_{j=1}^K  \mathcal{N}\left[r,\{i,j\}\right] v_j I_2(i, j)
\right], \label{eq:ODE_v_dropout}
\end{align}
where 
\begin{align}
J_2 &\equiv  \frac{2}{\pi} \left( 1 + c_{11} + c_{22} + c_{11} c_{22} - c_{12}^2 \right)^{-1/2}, \\
I_2 &\equiv \frac{1}{\pi} \arcsin\left(\frac{c_{12}}{\sqrt{1 + c_{11}} \sqrt{1 + c_{12}}}\right),  \\
I_3 &\equiv \frac{2}{\pi} \frac{1}{\sqrt{\Lambda_3}} \frac{c_{23}(1 + c_{11}) - c_{12} c_{13}}{1 + c_{11}}, \\
I_4 &\equiv  \frac{4}{\pi^2} \frac{1}{\sqrt{\Lambda_4}} \arcsin\left(\frac{\Lambda_0}{\sqrt{\Lambda_1 \Lambda_2}}\right), 
\end{align}
and
\begin{align}
\Lambda_4 &= (1 + c_{11})(1 + c_{22}) - c_{12}^2, \\
\Lambda_3 &=(1+c_{11})*(1+c_{33})-c_{13}^2\,,\\
\Lambda_0 &= \Lambda_4 c_{34} - c_{23} c_{24}(1 + c_{11}) - c_{13} c_{14}(1 + c_{22}) +   c_{12} c_{13} c_{24}+c_{12} c_{14} c_{23} , \\
\Lambda_1 &= \Lambda_4 (1 + c_{33}) - c_{23}^2(1 + c_{11}) - c_{13}^2(1 + c_{22}) + 2 c_{12} c_{13} c_{23},\\
\Lambda_2 &= \Lambda_4 (1 + c_{44}) - c_{24}^2(1 + c_{11}) - c_{14}^2(1 + c_{22}) + 2 c_{12} c_{14} c_{24}.
\end{align}
Note that we are using the indices $i,j,k,l$ to indicate the student's nodes, while $n,m$ indicate the teacher's nodes. When using the functions $I_2$, $I_3$, and $I_4$, we adopt the compact notation used in Ref.~\cite{goldt2019dynamics}. For instance, $I(i,n)$ takes as input the correlation matrix of the preactivations corresponding to the indices $i$ and $n$, i.e., $\lambda_i=w_i x/\sqrt{N}$ and $\lambda^*_n=w^*_n x/\sqrt{N}$. For this example, the correlation matrix would be
\begin{equation}
    C=\begin{pmatrix}
c_{11} & c_{12} \\
c_{21} & c_{22}
\end{pmatrix}=\begin{pmatrix}
\langle\lambda_i\lambda_i \rangle & \langle \lambda_i \lambda^*_n\rangle \\
\langle \lambda^*_n \lambda_i\rangle& \langle \lambda^*_n\lambda^*_n\rangle\end{pmatrix}=\begin{pmatrix}
Q_{ii} & R_{in} \\
R_{in} & T_{nn}
\end{pmatrix}
\,.
\end{equation}
Finally, the generalization error can be written as
\begin{align}
    \epsilon_g &= \frac{r_f^2}{\pi} \sum_{i,k} v_i v_k \arcsin\left(\frac{Q_{ik}}{\sqrt{1 + Q_{ii}} \sqrt{1 + Q_{kk}}}\right)
+ \frac{1}{\pi} \sum_{n,m} v_n^* v_m^* \arcsin\left(\frac{T_{nm}}{\sqrt{1 + T_{nn}} \sqrt{1 + T_{mm}}}\right)\\
&- \frac{2r_f}{\pi} \sum_{i,n} v_i v_n^* \arcsin\left(\frac{R_{in}}{\sqrt{1 + Q_{ii}} \sqrt{1 + T_{nn}}}\right).\label{eq:gen_err_appendix}
\end{align}

We report here the expression of the fixed-point equations $f_R=0$ and $f_Q=0$ in the small-$\eta$ limit. These can be obtained by neglecting the terms in $\eta^2$ in Eqs.~\eqref{eq:ODE_R_dropout}-\eqref{eq:ODE_Q_dropout}, yielding
\begin{align}
& 
    \sum_{m=1}^M \mathcal{N}\left[r,\{i\}\right]I_3(i, n, m) - \sum_{j=1}^K  \mathcal{N}\left[r,\{i,j\}\right] I_3(i, n, j) = 0\,, \label{eq:leadingR}\\
&   
    \sum_{m=1}^M \mathcal{N}\left[r,\{i\}\right] I_3(i, k, m) - \sum_{j=1}^K \mathcal{N}\left[r,\{i,j\}\right] I_3(i, k, j) \notag \\
&\quad + \sum_{m=1}^M \mathcal{N}\left[r,\{k\}\right] I_3(k, i, m) - \sum_{j=1}^K \mathcal{N}\left[r,\{k,j\}\right] I_3(k, i, j) = 0\,. \label{eq:leadingQ}
\end{align}
Additionally, assuming the specialization ansatz in Eq.~\eqref{eq:ansatz_specialization}, the ODEs can be written, in the case $K=M\geq2$ and in the small-$\eta$ limit, as
\begin{align}
\label{eq:special_eqR}
   \frac{ \mathrm{d}R}{\mathrm{d}\alpha}&=\eta r I_3\left(\begin{array}{ccc}Q & R &R\\ R& 1 &1\\ R &1 &1
   \end{array}\right)+\eta r (M-1) I_3\left(\begin{array}{ccc}Q & R &S\\ R& 1 &0\\ S& 0 &1\end{array}\right)-\eta r I_3\left(\begin{array}{ccc}Q & R &Q\\ R& 1 &R\\ Q &R  &Q
   \end{array}\right)\\
   & -\eta r^2 (M-1) I_3\left(\begin{array}{ccc}Q & R &C\\ R& 1 &S\\ C &S  &Q
   \end{array}\right)\,.\nonumber
\end{align}
\begin{align}
\label{eq:special_eqS}
   \frac{ \mathrm{d}S}{\mathrm{d}\alpha}&=\eta r I_3\left(\begin{array}{ccc}Q & S &R\\ S& 1 &0\\ R &0 &1
   \end{array}\right)+\eta r I_3\left(\begin{array}{ccc}Q & S &S\\ S& 1 &1\\ S& 1 &1\end{array}\right)+\eta (M-2) r I_3\left(\begin{array}{ccc}Q & S &S\\ S& 1 &0\\ S &0  &1
   \end{array}\right)\\ -&\eta r I_3\left(\begin{array}{ccc}Q & S &Q\\ S& 1 &S\\ Q &S &Q
   \end{array}\right)-\eta r^2 I_3\left(\begin{array}{ccc}Q & S &C\\ S& 1 &R\\ C& R &Q\end{array}\right)-\eta (M-2) r^2 I_3\left(\begin{array}{ccc}Q & S &C\\ S& 1 &S\\ C &S  &Q
   \end{array}\right) \,.\nonumber
\end{align}
\begin{align}
\label{eq:special_eqQ}
   \frac{ \mathrm{d}Q}{\mathrm{d}\alpha}&=2\eta r I_3\left(\begin{array}{ccc}Q & Q &R\\ Q& Q &R\\ R &R &1
   \end{array}\right)+2\eta r(M-1) I_3\left(\begin{array}{ccc}Q & Q &S\\ Q& Q &S\\ S& S &1\end{array}\right)\\ -&2\eta r I_3\left(\begin{array}{ccc}Q & Q &Q\\ Q& Q &Q\\ Q &Q &Q
   \end{array}\right)-2\eta r^2 (M-1)I_3\left(\begin{array}{ccc}Q & Q &C\\ Q& Q &C\\ C& C &Q\end{array}\right) \,.\nonumber
\end{align}
\begin{align}
\label{eq:special_eqC}
   \frac{ \mathrm{d}C}{\mathrm{d}\alpha}&=2\eta r I_3\left(\begin{array}{ccc}Q & C &R\\ C& Q &S\\ R &S &1
   \end{array}\right)+2\eta r I_3\left(\begin{array}{ccc}Q & C&S\\ C& Q &R\\ S& R &1\end{array}\right)+2\eta (M-2) r I_3\left(\begin{array}{ccc}Q & C &S\\ C& Q &S\\ S &S  &1
   \end{array}\right)\\ -&\eta r (r+1) I_3\left(\begin{array}{ccc}Q & C &Q\\ C& Q &C\\ Q &C &Q
   \end{array}\right)-\eta r(r+1) I_3\left(\begin{array}{ccc}Q & C &C\\ C& Q &Q\\ C& Q &Q\end{array}\right)-2\eta (M-2) r^2 I_3\left(\begin{array}{ccc}Q & C &C\\ C& Q &C\\ C &C  &Q
   \end{array}\right) \,.\nonumber
\end{align}

\end{document}